\documentclass[10pt,twocolumn,letterpaper]{article}
\usepackage[pagenumbers]{cvpr}
\usepackage{array}
\definecolor{heatRed}{HTML}{FC8081}
\definecolor{heatOrg}{HTML}{FABE83}
\definecolor{heatYel}{HTML}{FBF69F}
\definecolor{cvprblue}{rgb}{0.21,0.49,0.74}
\usepackage[breaklinks,colorlinks,allcolors=cvprblue]{hyperref}
\def\confName{CVPR}
\def\confYear{2027}
\title{CutAgent: Creative-Gene-Driven Planning and Experience-Guided Self-Refinement for High-CTR E-commerce Advertising Video Creation}
\author{
  \textbf{Xuyue Huang}$^{1,2,*}$\quad\textbf{Guanbo Huang}$^{1,2,*}$\quad\textbf{Jiasong Feng}$^{3,*}$\quad\textbf{Run Ling}$^{1,*}$\quad\textbf{Ao Ma}$^{1,\dagger,\ddagger}$\quad\textbf{Xi Zhao}$^{1}$\\
  \textbf{Yaoyu Li}$^{1}$\quad\textbf{Jie Gao}$^{1}$\quad\textbf{Ke Cao}$^{1}$\quad\textbf{Wei Feng}$^{1}$\quad\textbf{Zhanjie Zhang}$^{4,\ddagger}$\quad\textbf{Hao Wang}$^{1}$\quad\textbf{Jinghe Hu}$^{1}$\\[2pt]
  $^{1}$JD.com\qquad$^{2}$Tsinghua University\qquad$^{3}$Tianjin University\qquad$^{4}$Zhejiang University\\[2pt]
  \texttt{\{huang-xy24, hgb24\}@mails.tsinghua.edu.cn}\quad\texttt{maao.8@jd.com}
}
\begin{document}
\maketitle
\begingroup
  \renewcommand{\thefootnote}{\fnsymbol{footnote}}
  \footnotetext[1]{Equal contribution.}
  \footnotetext[2]{Project leader.}
  \footnotetext[3]{Corresponding authors.}
\endgroup

\begin{abstract}
E-commerce advertising video creation requires both high-level creative
planning and reliable multi-modal production. However, existing methods
often map product attributes directly to scripts or editing decisions
without explicitly modeling advertising strategies, narrative structures,
and production styles. They also have limited ability to reuse past
production failures.
To address these limitations, we propose \textbf{CutAgent}, a dual-agent
framework that formulates advertising video creation as a
Plan--Execute--Self-Evolve process. CutAgent comprises two collaborative
agents, \textbf{GeneAgent} and \textbf{VideoAgent}, connected by a
structured creative gene that bridges product understanding and video
production. GeneAgent maps multi-modal product inputs---textual attributes
together with product images---to creative genes that specify narrative
units and production rules. To train GeneAgent, we construct
\textbf{CutAgent-70K} with approximately 70K samples of product
attributes/images, reverse-extracted creative genes, cleaned clip pools,
and real-world CTR, and apply high-CTR supervised fine-tuning followed by
CTR-guided GRPO. Guided by the generated gene, VideoAgent retrieves
category-then-gene failure--correction experiences from a long-term
memory as constraints for shot-script planning, then coordinates material
retrieval, narration, subtitles, music, and timeline
rendering. Independently, a reflection loop diagnoses quality defects,
revises deficient production decisions, and writes verified corrections
back to memory for later retrieval. Experiments on
\textbf{CutAgent-Eval} show that CutAgent outperforms representative
baselines in e-commerce advertising video creation, achieving the highest
online CTR of \textbf{2.21\%}.

\end{abstract}


\section{Introduction}

\begin{figure*}[t]
  \centering
  \includegraphics[width=\linewidth]{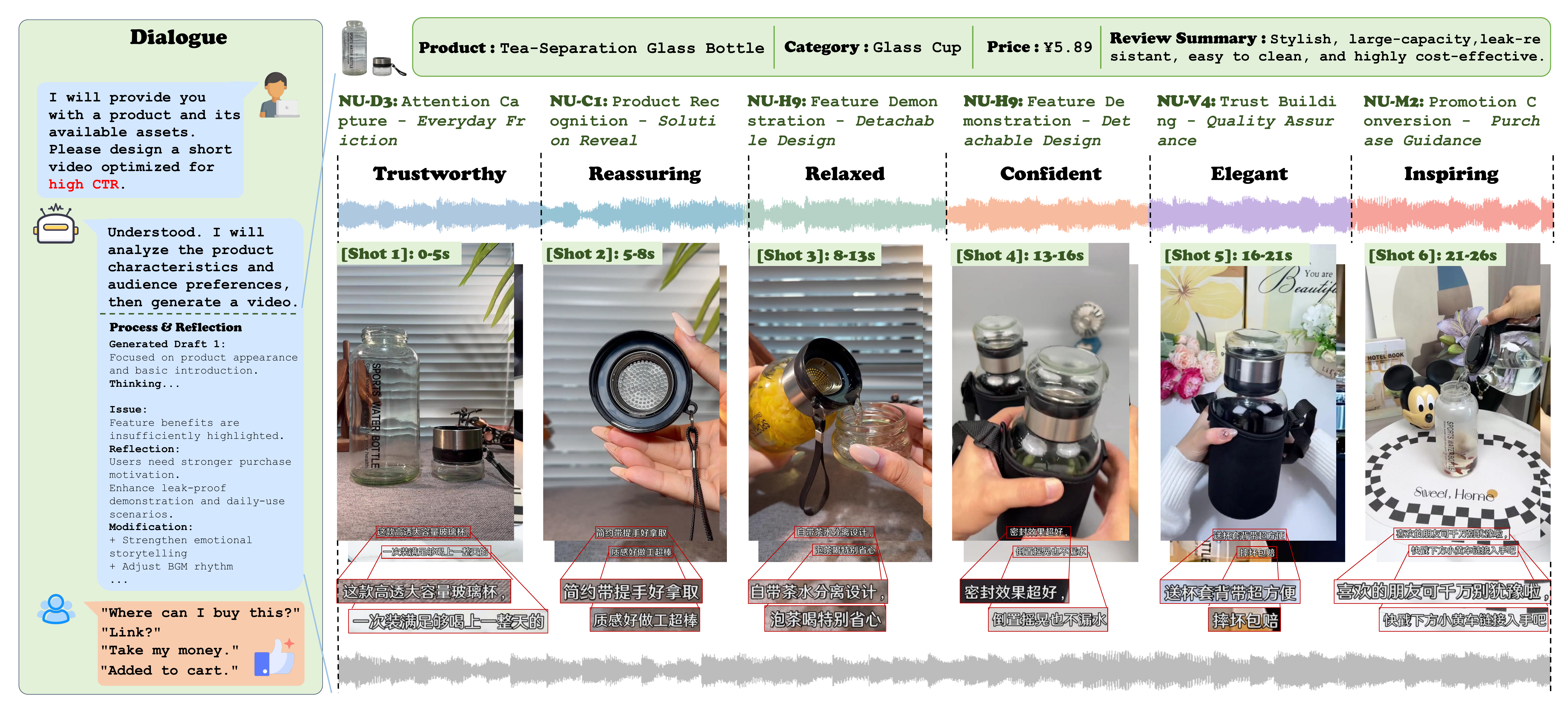}
\caption{
Illustration of CutAgent on a tea-separation glass bottle advertisement.
Given product information and assets, CutAgent generates a creative-gene-guided shot sequence covering attention capture, product recognition, feature demonstration, trust building, and purchase guidance.
The reflection loop diagnoses weak feature emphasis in the initial draft and refines the video with stronger demonstrations, daily-use scenarios, and improved audio rhythm.
} 
  \label{fig:teaser}
\end{figure*}

On modern e-commerce platforms, short-form advertising videos have become an important medium for product promotion and user engagement~\cite{qian2025vc,liu2026vads,zhou2026autocut}. Compared with static product pages that primarily present specifications and prices, short videos can demonstrate product usage, communicate selling points, and encourage immediate user interaction within a limited duration. In this setting, click-through rate (CTR) provides a direct behavioral signal of whether an advertisement successfully captures attention, conveys product value, and motivates users to engage. Consequently, platforms and merchants increasingly demand scalable methods for producing advertising videos with strong conversion
potential.

Producing such videos, however, is far from a single generation step. An effective advertisement must decide which selling points to emphasize, in what narrative order they should appear, and how they should be realized through materials, scripts, voice-over, subtitles and background music. Professional editors routinely iterate over these intertwined decisions, yet the cost of manual production grows rapidly with the number of SKUs, seasonal campaigns, and category variations. This tension between conversion-oriented creative quality
and production scalability motivates automated high-CTR advertising video creation.

While automated video production has rapidly advanced from template-based retrieval~\cite{bain2021frozen} to LLM-guided planning~\cite{maaz2024video,lin2023videodirectorgpt} and sophisticated agentic workflows~\cite{zhou2026autocut,zhao2026cutclaw,firered2026openstoryline}, generating high-CTR advertising videos remains under-addressed along two critical dimensions. First, existing pipelines map product attributes directly to low-level editing decisions without a globally coherent advertising strategy. Because creative intent is entangled with execution, it becomes exceedingly difficult to optimize the planning process using commercial feedback like CTR. Second, advertising video production is inherently error-prone. Current systems rarely diagnose failures—such as missing selling points or mismatched audio—at the decision level, nor do they reuse past failure experiences to prevent the same mistakes on subsequent products.


To close this gap, we propose \textbf{CutAgent}, a dual-agent framework that formulates e-commerce advertising video creation as a \textbf{Plan--Execute--Self-Evolve} paradigm. By introducing an explicit \emph{creative gene} (i.e., ordered narrative units and production rules) as an intermediate representation, CutAgent decouples conversion-oriented creative planning from low-level production operations. This design ensures that the generated videos adhere to a globally coherent advertising strategy while making the pipeline explicitly controllable, amenable to CTR-guided learning, and capable of continuous self-refinement. 

Specifically, CutAgent operates through two synergistic agents to map multi-modal product inputs to high-CTR video outputs. In the planning stage, the \textbf{GeneAgent} processes product attributes to generate the creative gene. To ensure commercial effectiveness, it is trained on our newly constructed \textbf{CutAgent-70K} dataset using supervised fine-tuning~\cite{ouyang2022training} on top-tier data, followed by CTR-guided Group Relative Policy Optimization (GRPO)~\cite{shao2024deepseekmath} with a surrogate reward model~\cite{ke2017lightgbm}. Subsequently, the \textbf{VideoAgent} handles execution and self-evolution. Guided by the generated gene and a long-term memory of past failure--correction experiences, it coordinates script expansion and video rendering. Crucially, VideoAgent employs an independent post-production reflection loop to diagnose deficient decisions, revise affected components, and write verified corrections back into memory for future tasks. Extensive experiments on our \textbf{CutAgent-Eval} benchmark demonstrate that CutAgent successfully automates and optimizes ad generation, outperforming representative methods and achieving a highly competitive online CTR of 2.21\%.

Our main contributions are summarized as follows:
\begin{itemize}
    \item We construct \textbf{CutAgent-70K}, a large-scale dataset connecting multi-modal product attributes and images, reverse-extracted creative genes, cleaned video-clip pools, and real-world CTR signals, together with \textbf{CutAgent-Eval} for comprehensive evaluation of advertising video planning and production.

    \item We introduce \textbf{GeneAgent}, a conversion-oriented creative planning agent that learns product-conditioned creative genes from multi-modal inputs of product attributes and product images through high-CTR supervised fine-tuning and CTR-guided GRPO.

    \item We propose \textbf{VideoAgent}, which translates creative genes into complete advertising videos under category-then-gene constraints retrieved from a long-term memory, and independently performs six-dimensional diagnosis with reflection-based iterative refinement, writing verified corrections back to memory.

    \item Extensive experiments demonstrate that the resulting \textbf{CutAgent} framework consistently outperforms representative open-source and proprietary methods on CutAgent-Eval.
\end{itemize}

\section{Related Work}
\label{sec:related}
\subsection{E-commerce Video Creation}

\begin{figure*}[t]
  \centering
  \includegraphics[width=\linewidth]{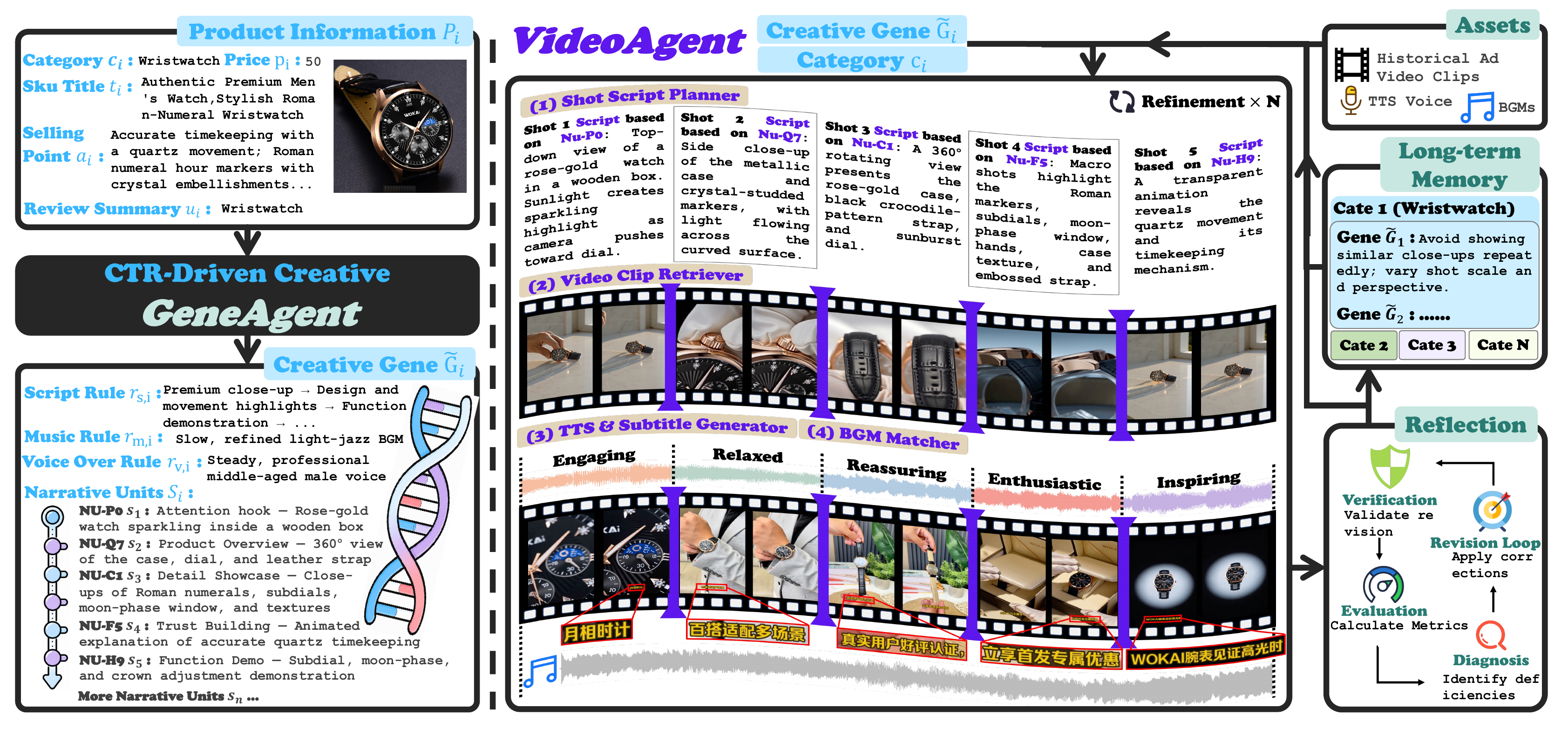}
  \caption{
  Overview of CutAgent. GeneAgent takes product attributes and product
  images as multi-modal input $P_i$ and maps them to a creative gene
  (narrative units with voice-over, music and script rules). Conditioned on the gene, VideoAgent first
  retrieves category-then-gene failure--correction experiences from a
  long-term memory $\mathcal{E}$, then executes creative-gene-guided shot
  planning, video-clip retrieval, TTS and subtitle generation, BGM matching, and timeline rendering. Independently, a
  reflection loop performs evaluation, diagnosis, revision, and
  verification after rendering, and writes verified experiences back to
  $\mathcal{E}$ for later retrieval.
  }
  \label{fig:framework}
\end{figure*}

Recent studies have explored agentic frameworks for automating complex video editing workflows~\cite{wang2024lave,yu2026aurora}. CutClaw~\cite{zhao2026cutclaw} employs hierarchical audiovisual decomposition together with Playwriter, Editor, and Reviewer agents to transform hours-long raw footage into instruction-following, music-synchronized short videos. FireRed-OpenStoryline~\cite{firered2026openstoryline} adopts a tool-oriented architecture in which language models coordinate media retrieval, clip understanding, script generation, timeline planning, and rendering.

For e-commerce video creation, AutoCut~\cite{zhou2026autocut} discretizes visual and audio representations into multi-modal tokens and fine-tunes a multi-modal language model for clip selection, clip ordering, script generation, and background music selection. Although AutoCut uses click-through rate and like rate to select high-engagement training videos, these signals are used for data curation rather than as explicit optimization objectives. Existing methods~\cite{zhao2026cutclaw,firered2026openstoryline,zhou2026autocut} therefore lack a unified mechanism that combines conversion-oriented planning, post-production reflection, and reusable correction experience across production tasks. CutAgent addresses this gap by separating creative-gene planning from agentic video production and accumulating verified failure--correction experiences in a category-then-gene long-term memory.

\subsection{Planning and Self-Evolving Agents}

Large language model (LLM) agents have emerged as a general paradigm
for solving complex tasks through planning, tool use, and interaction
with external environments~\cite{wang2024survey}. Among representative
approaches, ReAct~\cite{yao2023react} interleaves reasoning with
environmental actions, allowing an agent to update its decisions based
on intermediate observations, while
HuggingGPT~\cite{shen2023hugginggpt} uses a language model for task
planning, model selection, subtask execution, and result aggregation.

Beyond single-task execution, self-evolving agents seek to improve
their future behavior by acquiring, refining, and reusing experiences
from previous interactions~\cite{gao2025survey}. Among
representative approaches, Reflexion~\cite{shinn2023reflexion}
transforms evaluation feedback into verbal reflections stored in
episodic memory, whereas ExpeL~\cite{zhao2024expel} extracts reusable
insights and successful experiences from historical trajectories.
However, these approaches mainly target general reasoning and tool-use
tasks, without jointly considering structured creative planning,
multi-modal production, outcome-driven evaluation, and the reuse of
dimension-specific production experiences. CutAgent extends this
paradigm to e-commerce advertising video creation by separating
creative-gene planning from video execution and accumulating verified
failure--correction experiences in a category-then-gene long-term memory.

\section{Method}

\subsection{Overview}

Fig.~\ref{fig:framework} illustrates the overall CutAgent pipeline.
The input of CutAgent is a multi-modal product representation $P_i$ that
contains both textual product attributes. Conditioned on $P_i$, \textbf{GeneAgent} performs
conversion-oriented planning and outputs a structured creative gene
$\widetilde{G}_i$. As illustrated in the figure, the gene comprises an ordered sequence of narrative units and corresponding audiovisual production rules. This gene answers \emph{what kind of advertisement should be
created} before any clip is selected.


Conditioned on the product and creative gene, \textbf{VideoAgent} executes the video production. Crucially, as illustrated in Figure~\ref{fig:framework}, this process is augmented by two independent modules: a \emph{memory-read} step before script expansion, and a \emph{reflection-write} step after rendering. \textbf{First}, given a new gene $\widetilde{G}_i$, VideoAgent retrieves failure--correction experiences from long-term memory. \textbf{Second}, the agent executes the script by retrieving visual clips from a product-specific pool $\mathcal{M}_i$, synthesizing audio and subtitles and rendering the final timeline. \textbf{Finally}, an independent reflection loop evaluates the rendered video, diagnoses deficient decisions, and writes newly verified failure--correction pairs back into $\mathcal{E}$ to prevent recurring mistakes in future tasks.

CutAgent therefore follows
\textbf{Plan--Execute--Self-Evolve}: GeneAgent plans the creative gene,
VideoAgent executes multi-modal production under memory-retrieved
constraints, and reflection enables non-parametric self-evolution by
updating $\mathcal{E}$ across products.

\begin{figure*}[t]
  \centering
  \includegraphics[width=\linewidth]{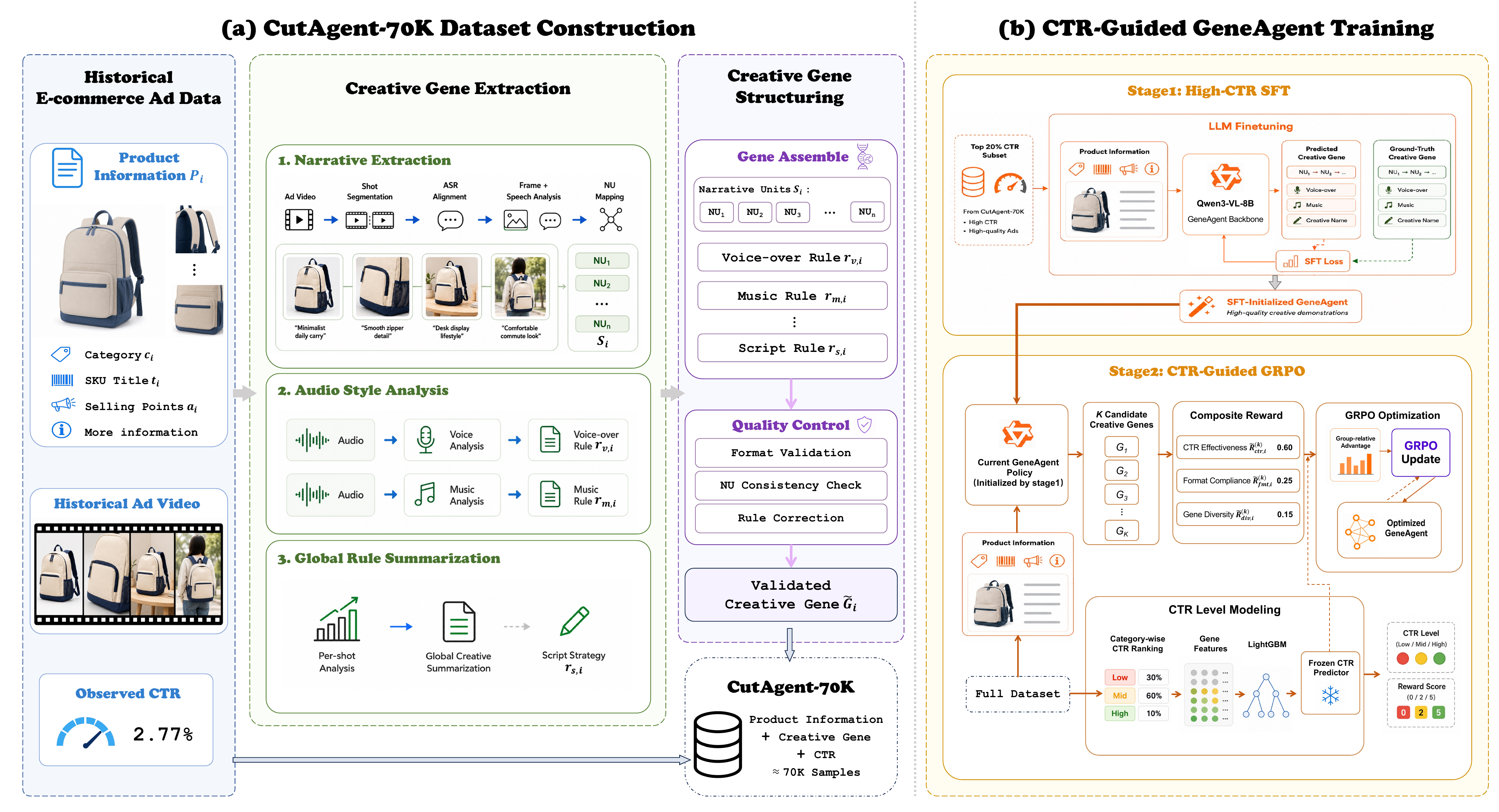}
  \caption{
  Construction of CutAgent-70K and CTR-guided GeneAgent training.
  (a)~For each product, historical ads are cleaned into a clip pool and
  processed by reverse gene extraction for CutAgent-70K
  (see Fig.~\ref{fig:prompts}(C1--C4)), followed by quality control; the
  resulting sample contains product attributes and images $P_i$, gene
  $G_i$, cleaned clips $\mathcal{M}_i$, and CTR $y_i$.
  (b)~GeneAgent (Qwen3-VL-8B) is first supervised on high-CTR samples,
  then optimized by GRPO: $K{=}8$ candidate genes are sampled and scored
  by a composite reward (CTR effectiveness 0.60, format compliance 0.25,
  gene diversity 0.15) using a frozen LightGBM CTR-level predictor as
  the training-time surrogate reward (not for evaluation).
  }
  \label{fig:data_pipeline}
\end{figure*}

\subsection{GeneAgent}

\subsubsection{Creative Gene Representation}
For the $i$-th product, the multi-modal input is
\begin{equation}
P_i =
\left\{
c_i,t_i,p_i,a_i,
I_i^{\mathrm{main}},
I_i^{\mathrm{detail}},
u_i
\right\},
\end{equation}
where $c_i$, $t_i$, $p_i$, and $a_i$ denote textual product attributes
(category, SKU title, price, and selling points),
$I_i^{\mathrm{main}}$ and $I_i^{\mathrm{detail}}$ are the main image and
detail images of the product, and $u_i$ is the review summary. Thus
$P_i$ always couples product attributes with product images, which is
necessary for generating shot plans that remain visually aligned with
the target product.

GeneAgent is a vision--language model that consumes the complete
multi-modal product representation $P_i$. Following the creative-gene
prompt in Fig.~\ref{fig:prompts}(A),
it predicts a structured creative gene:
\begingroup
\small
\begin{equation}
\widetilde{G}_i
=
\mathrm{GeneAgent}(P_i)
=
\left(
\widetilde{S}_i,
\widetilde{r}_{v,i},
\widetilde{r}_{m,i},
\widetilde{r}_{s,i}
\right),
\end{equation}
where
\begin{equation}
\widetilde{S}_i
=
\left[
\widetilde{s}_{i,1},\ldots,\widetilde{s}_{i,n_i}
\right],
\widetilde{s}_{i,j}
=
\left(
\widetilde{u}_{i,j},
\widetilde{d}_{i,j}
\right),
\widetilde{u}_{i,j}\in\mathcal{NU}.
\end{equation}
\endgroup
Here, $\widetilde{S}_i$ is an ordered sequence of narrative units.
Each narrative unit consists of a discrete type
$\widetilde{u}_{i,j}$ from the predefined vocabulary $\mathcal{NU}$
and a product-conditioned textual realization
$\widetilde{d}_{i,j}$ specifying how the corresponding narrative
function should be instantiated for the target product (the complete 27-unit vocabulary is listed in Table~\ref{tab:nu_registry}).

The remaining fields denote the voice-over, music and script rules, respectively. The creative gene serves as an
explicit interface between CTR-oriented creative planning and
downstream video production.

\subsubsection{Dataset Construction}
\label{sec:dataset}
We construct CutAgent-70K for GeneAgent training and VideoAgent
material retrieval, and independently curate CutAgent-Eval for
final evaluation (see Fig.~\ref{fig:data_pipeline}(a)).

CutAgent-70K contains approximately 70K product-level samples covering
43 product categories organized into 8 portfolios
(see Fig.~\ref{fig:dataset_sta}). For each product, we collect
multi-modal product information $P_i$, including textual attributes and
main/detail images, together with a product-specific material pool
$\mathcal{M}_i$ containing approximately 90--120 cleaned video clips.
We further recover the corresponding creative gene $G_i$ through a
four-stage reverse-extraction pipeline
(see Fig.~\ref{fig:prompts}C). The resulting dataset is
\begin{equation}
\mathcal{D}_{70K}
=
\left\{
(P_i,G_i,\mathcal{M}_i,y_i)
\right\}_{i=1}^{N},
\quad
N\approx70\mathrm{K},
\end{equation}
where $y_i=C_i/E_i$ is the observed CTR. GeneAgent training uses
$(P_i,G_i,y_i)$; VideoAgent retrieves shots from $\mathcal{M}_i$ at
production time.

Cleaning uses ffmpeg scene
detection, removes clips with on-camera humans or under 2\,s, blurs
subtitles/watermarks, and deduplicates by perceptual hash. Reverse extraction uses
Qwen3.5-Omni-Plus-Realtime for voice/BGM attributes and
Doubao-Seed-2.0-pro~\cite{seed2026seed2} for shot analysis and gene
reconstruction.

\subsubsection{CutAgent-Eval}
CutAgent-Eval is an independent 5K-product set from the same
43-category taxonomy, unused in SFT, reward-model training, or GRPO;
all methods share its product inputs and standardized clip pools.

\subsubsection{GeneAgent Optimization}
As illustrated in Fig.~\ref{fig:data_pipeline}(b), GeneAgent is optimized
in two stages. High-CTR supervised fine-tuning first establishes the
product-to-gene mapping, after which CTR-guided GRPO improves the
advertising potential of generated creative genes.

\subsubsection{High-CTR Supervised Fine-Tuning}
\label{sec:sft}
We select the top 20\% samples ranked by observed CTR from CutAgent-70K
as high-quality creative demonstrations:
\begin{equation}
\mathcal{D}_{\mathrm{SFT}}
=
\mathrm{Top}_{20\%}
\left(
\mathcal{D}_{70K};
\mathrm{CTR}
\right).
\end{equation}

We employ Qwen3-VL-8B-Instruct~\cite{bai2025qwen3} as the vision--language
backbone of GeneAgent. For each training sample, the model input is the
full multi-modal product representation $P_i$, i.e., textual product
attributes together with the main and detail product images, under the
creative-gene prompt (see Fig.~\ref{fig:prompts}A). Conditioned
on $P_i$, GeneAgent autoregressively generates the serialized creative
gene $\widetilde{G}_i=[g_{i,1},\ldots,g_{i,T_i}]$:
\begin{equation}
p_{\theta}
\left(
\widetilde{G}_i\mid P_i
\right)
=
\prod_{t=1}^{T_i}
p_{\theta}
\left(
g_{i,t}
\mid
P_i,g_{i,<t}
\right).
\end{equation}

The supervised fine-tuning objective is
\begin{equation}
\begin{aligned}
\mathcal{L}_{\mathrm{SFT}}
&=
-\frac{1}{|\mathcal{D}_{\mathrm{SFT}}|}
\sum_{(P_i,G_i)\in\mathcal{D}_{\mathrm{SFT}}}
\sum_{t=1}^{T_i}
m_{i,t}
\\
&\qquad\log
p_{\theta}
\left(
g_{i,t}
\mid
P_i,g_{i,<t}
\right),
\end{aligned}
\end{equation}
where $m_{i,t}$ masks the instruction tokens, textual product-attribute
tokens, and visual tokens of the product images, so that the loss is
computed only over the target gene tokens. We freeze the visual encoder
and vision--language alignment module and optimize the LoRA parameters
of the language model. The resulting SFT model initializes GRPO and
serves as fixed reference policy for KL regularization.

\subsubsection{CTR-Level Reward Modeling}
\label{sec:ctr_reward}
Raw CTR distributions vary substantially across categories. We rank
historical advertisements within each category by observed CTR and
divide them into Low (bottom 30\%), Mid (30--90\%), and High (top 10\%)
levels. A frozen LightGBM~\cite{ke2017lightgbm} classifier is trained on
handcrafted features $\phi(P_i,G_i)\in\mathbb{R}^{d}$ (narrative-unit
sequences, gene-rule keywords, TF--IDF of textual fields, and product
attributes) to predict these CTR levels as a surrogate reward during
GRPO training only (see Sec.~\ref{app:ctr_reward_hparams}).

\subsubsection{CTR-Guided GRPO}
\label{sec:grpo}
For each product input $P_i$, the old policy samples $K{=}8$ candidate
creative genes
$\widehat{G}_i^{(k)}\sim\pi_{\theta_{\mathrm{old}}}(G\mid P_i)$. Each
candidate is scored by a composite reward combining three terms:
$
R_i^{(k)}
=
0.60\widetilde{R}_{\mathrm{ctr},i}^{(k)}
+
0.25R_{\mathrm{fmt},i}^{(k)}
+
0.15R_{\mathrm{div},i}^{(k)},
$
where $\widetilde{R}_{\mathrm{ctr}}$ is the normalized CTR-level reward
from the frozen surrogate, $R_{\mathrm{fmt}}$ penalizes format violations
(field completeness and serialization correctness), and $R_{\mathrm{div}}$
discourages repetitive candidates via Jaccard dissimilarity over
narrative-unit sets and production-rule fields.

The GRPO objective maximizes the clipped surrogate with KL regularization against the frozen SFT reference policy, as shown in Eq.~\eqref{eq:grpo}, where $\rho_{i,k,t}$ is the token-level importance ratio,
$\widehat{A}_i^{(k)}$ is the group-relative advantage computed from the
$K$ candidates of the same product, $\epsilon{=}0.2$ is the clipping
coefficient, and $\beta{=}0.04$ controls the KL penalty against the
fixed SFT model. Full derivations of the reward components, advantage
normalization, and KL term are provided in
Sec.~\ref{app:ctr_reward_hparams}.

\begin{equation}
\label{eq:grpo}
\begin{aligned}
\mathcal{L}_{\mathrm{GRPO}}(\theta)
&=
-\mathbb{E}
\Bigg[
\frac{1}{K}
\sum_{k=1}^{K}
\frac{1}{T_{i,k}}
\sum_{t=1}^{T_{i,k}}
\Bigg(
\\
&\qquad\min
\Big[
\rho_{i,k,t}(\theta)\widehat{A}_i^{(k)},\\
&\qquad
\mathrm{clip}
\left(
\rho_{i,k,t}(\theta),
1{-}\epsilon,
1{+}\epsilon
\right)
\widehat{A}_i^{(k)}
\Big]
\\
&\qquad-
\beta D_{\mathrm{KL}}^{i,k,t}
\Bigg)
\Bigg],
\end{aligned}
\end{equation}

\subsection{VideoAgent}

Given $P_i$ and $\widetilde{G}_i$, VideoAgent produces the advertising
video as follows (Fig.~\ref{fig:framework}). Before expanding the gene
into scripts, it retrieves category-then-gene constraints from the
long-term memory $\mathcal{E}$ (detailed under Long-Term Memory below).
It then executes shot-script planning and narration/subtitle adaptation, alongside clip retrieval, TTS synthesis, BGM matching, and timeline rendering.

\subsubsection{Creative-Gene-Guided Shot Planning}
Conditioned on the creative gene, and the retrieved constraints from \textbf{Long-Term Memory} and \textbf{Reflection}, VideoAgent expands the gene into an executable shot-by-shot script, yielding a storyboard
$
\mathcal{B}_i
=
[
b_{i,1},\ldots,b_{i,L_i}
]
$,
where each narrative unit maps to one representative shot with a visual
description, voice-over text, duration, and production requirements.
Narrative units determine shot goals and order; gene rules control
script style, voice-over and music; product
images keep shot descriptions appearance-consistent with the advertised
product; and $\mathcal{E}_{i}^{\mathrm{ret}}$ constrains error-prone
choices during script expansion.

\subsubsection{Video-Clip Retrieval, TTS, BGM, and Rendering}
VideoAgent retrieves clips from $\mathcal{M}_i$ with
Qwen3-VL-Embedding-8B~\cite{li2026qwen3};
CosyVoice~2~\cite{du2024cosyvoice} synthesizes narration; BGM combines Qwen3-Embedding-8B
matching with librosa tempo, energy, and spectral
cues~\cite{mcfee2015librosa}; ffmpeg renders the
timeline~\cite{tomar2006converting}.

\subsubsection{Reflection}

Although the creative gene provides a global production plan, residual
errors may still arise during execution. VideoAgent therefore performs a
four-stage reflection loop on the rendered video: it \emph{evaluates} the
video along six dimensions (NQ, SPC, VSC, MQA, MSS, TB) with temporally
grounded evidence, \emph{diagnoses} deficient decisions by localizing
defects to specific shots or production modules, \emph{revises} only the
affected components, and \emph{verifies} the revised video under the same
protocol until quality requirements are met or the maximum iteration count
is reached (see Sec.~\ref{app:reflection}). Verified failure--correction
experiences are written into the long-term memory $\mathcal{E}$ and later
retrieved by category-then-gene matching to constrain script production
for new genes.

\subsubsection{Long-Term Memory}
\label{sec:long_term_memory}

Long-term memory complements the reflection loop through a
read--write mechanism for production experience. It is pre-populated
with verified failure--correction experiences extracted from historical
production cases in CutAgent-70K. Before a creative gene is expanded
into shot scripts, VideoAgent reads relevant experiences from memory
as production constraints. After rendering, the reflection loop
diagnoses execution failures and writes newly verified corrections
back into memory for future tasks. To reuse
production experience across products without parameter updates,
VideoAgent maintains
$
\mathcal{E}
=
\left\{
\mathcal{E}^{(c)}
\right\}_{c\in\mathcal{C}},
$
where $\mathcal{C}$ is the set of 43 product categories and
$\mathcal{E}^{(c)}$ stores experiences for category $c$, disjoint from
the clip pool $\mathcal{M}_i$. Each entry
$e=(\widetilde{G}_{e},d_{e},m_{e},c_{e})$ stores a gene context
$\widetilde{G}_{e}$, an observed defect description $d_{e}$, the affected
production module $m_{e}$, and a verified correction $c_{e}$
(e.g., ``avoid repeated close-ups in the opening and insert a
trust-building unit'').

Retrieval is performed in two stages \emph{before} shot-script
generation. First, given the product category $c_i$, VideoAgent selects
the corresponding shard $\mathcal{E}^{(c_i)}$ so that only experiences
from the same category are candidates. Second, within
$\mathcal{E}^{(c_i)}$, it matches the current gene $\widetilde{G}_i$ to
stored gene contexts with BGE-M3~\cite{chen2024bge}: both genes are
serialized to text, embedded, and ranked by cosine similarity, yielding
the top-$K$ experiences
\begin{equation}
\begin{aligned}
\mathcal{E}_{i}^{\mathrm{ret}}
&=
\mathrm{TopK}
\Big(
\big\{
e\in\mathcal{E}^{(c_i)}
\big\},\;\\
&\qquad
\cos\!\big(
\mathrm{BGE\text{-}M3}(\widetilde{G}_i),\,
\mathrm{BGE\text{-}M3}(\widetilde{G}_{e})
\big)
\Big).
\end{aligned}
\end{equation}
The retrieved corrections are injected as explicit constraints that tell
the planner \emph{what this gene should attend to} when producing shot
scripts---for example, avoiding redundant close-ups, covering omitted
selling points, or preserving trust-building beats. Thus memory
retrieval is category-then-gene conditioned and independent of running
reflection on the \emph{current} video; reflection on prior videos is
what populates $\mathcal{E}$.

\begin{table*}[t]
\centering
\caption{
Quantitative comparison with creative-planning and advertising-video
creation baselines on CutAgent-Eval.
}
\label{tab:main_comparison}
\footnotesize
\setlength{\tabcolsep}{4.0pt}
\renewcommand{\arraystretch}{1.10}

\begin{tabular}{lcccccc|cc}
\toprule
&
\multicolumn{6}{c|}{Video Quality}
&
\multicolumn{1}{c}{Advertising Effectiveness}
&
\multicolumn{1}{c}{Human Evaluation}
\\
\cmidrule(lr){2-7}
\cmidrule(lr){8-8}
\cmidrule(lr){9-9}

Method
& NQ$\uparrow$
& SPC$\uparrow$
& VSC$\uparrow$
& MQA$\uparrow$
& MSS$\uparrow$
& TB$\uparrow$
& Online CTR (\%)$\uparrow$
& A/B Pref. (\%)$\uparrow$
\\

\midrule
\multicolumn{9}{l}{\textit{Creative-Gene-Driven Planning}} \\

Qwen3-VL-8B-Instruct
& 86.8
& 79.6
& 83.2
& 84.7
& 85.4
& 66.8
& 1.12
& 37.5
\\

Seed-2.0-Pro
& 94.1
& 82.7
& 84.1
& 85.8
& 85.9
& 70.6
& 1.38
& 40.8
\\

DeepSeek-V3
& 88.9
& 83.5
& 84.7
& 86.2
& 86.3
& 72.1
& 1.51
& 39.7
\\

Kimi-2.5
& 89.7
& 85.2
& 85.5
& 87.0
& 86.8
& 74.5
& 1.78
& 42.3
\\

Gemini-3-Pro
& 90.1
& 86.1
& 85.9
& 87.5
& 87.1
& 76.2
& 1.03
& 41.2
\\

ChatGPT-5.6-Sol
& 87.8
& 88.0
& 86.5
& 88.2
& 87.5
& 78.6
& 1.91
& 44.1
\\

\midrule
\multicolumn{9}{l}{\textit{Advertising Video Creation}} \\

AutoCut
& 87.4
& 75.3
& 82.6
& 84.4
& 85.8
& 62.3
& 1.98
& 34.6
\\

FireRed-OpenStoryline
& 90.2
& 71.2
& 80.6
& 82.1
& \textbf{90.2}
& 68.2
& 1.82
& 36.9
\\

\midrule

\textbf{Ours (CutAgent)}
& \textbf{94.9}
& \textbf{90.9}
& \textbf{87.4}
& \textbf{89.6}
& 88.5
& \textbf{81.3}
& \textbf{2.21}
& \textbf{60.4}
\\

\bottomrule
\end{tabular}
\end{table*}

\section{Experiments}

\subsection{Implementation Details}
GeneAgent uses Qwen3-VL-8B-Instruct. We freeze the visual encoder and
vision--language alignment module, and apply LoRA to all linear layers
with rank 16, scaling factor 32, and dropout 0.05. SFT trains on the top
20\% CTR samples for three epochs with learning rate
$5\times10^{-5}$, batch size 32, and sequence length 4{,}096. GRPO then
uses all CutAgent-70K for one epoch with $K{=}8$, learning rate
$1\times10^{-6}$, clipping coefficient 0.2, and KL coefficient 0.04.
The reward model uses 31 leaves and 500 boosting rounds. The loss is
computed only over creative-gene tokens.

\subsection{Baselines and Evaluation Protocol}
To evaluate high-CTR e-commerce advertising video creation and the
effectiveness of creative-gene-driven planning, we compare CutAgent with
the end-to-end advertising video creation method
AutoCut~\cite{zhou2026autocut} and the agentic editing system
FireRed-OpenStoryline~\cite{firered2026openstoryline}. FireRed-OpenStoryline is an
open agentic editor that consumes user media and natural-language
directives to search/organize clips, generate scripts, recommend
music/voiceover, and render timelines; given the same product materials
and an advertising-oriented prompt, it is a fair tool-orchestration
baseline for end-to-end video production. To isolate gene
generation quality, we further evaluate creative genes produced by
Qwen3-VL-8B-Instruct~\cite{bai2025qwen3},
Seed-2.0-Pro~\cite{seed2026seed2}, DeepSeek-V3~\cite{liu2024deepseek},
Kimi-2.5~\cite{team2026kimi}, Gemini-3-Pro~\cite{team2023gemini}, and
ChatGPT-5.6-Sol~\cite{achiam2023gpt}, and convert all genes with the same
VideoAgent. All methods receive the same multi-modal product inputs (attributes and
images) and material pools on CutAgent-Eval and are assessed under an
identical evaluation protocol.

\subsection{Experimental Results}
\begin{table*}[t]
\scriptsize
\centering
\caption{
Ablation of SFT, GRPO, reflection, and long-term memory on
CutAgent-Eval.
}
\label{tab:cutagent_ablation}
\setlength{\tabcolsep}{2.8pt}
\renewcommand{\arraystretch}{1.12}

\begin{tabular}{lcccc|cccccc|cc}
\toprule
&
\multicolumn{4}{c|}{Configuration}
&
\multicolumn{6}{c|}{Video Quality}
&
\multicolumn{1}{c}{Advertising Effectiveness}
&
\multicolumn{1}{c}{Human Evaluation}
\\
\cmidrule(lr){2-5}
\cmidrule(lr){6-11}
\cmidrule(lr){12-12}
\cmidrule(lr){13-13}

Variant
& SFT
& GRPO
& Reflection
& Memory
& NQ$\uparrow$
& SPC$\uparrow$
& VSC$\uparrow$
& MQA$\uparrow$
& MSS$\uparrow$
& TB$\uparrow$
& Online CTR (\%)$\uparrow$
& A/B Pref. (\%)$\uparrow$
\\

\midrule

Base
&  &  &  &
& 86.8
& 79.6
& 83.2
& 84.7
& 85.4
& 66.8
& 1.12
& 37.5
\\

SFT
& \checkmark &  &  &
& 92.0
& 87.3
& 85.4
& 87.1
& 87.2
& 76.2
& 1.76
& 54.2
\\

SFT + Reflection
& \checkmark &  & \checkmark &
& 94.6
& 90.1
& 86.7
& 88.9
& 88.1
& \textbf{81.5}
& 1.89
& 59.1
\\

SFT + GRPO
& \checkmark & \checkmark &  &
& 93.0
& 88.8
& 87.3
& 88.0
& \textbf{88.7}
& 78.4
& 2.05
& 57.9
\\

SFT + GRPO + Reflection
& \checkmark
& \checkmark
& \checkmark
&
& 94.7
& 90.7
& 87.2
& \textbf{89.7}
& 88.3
& 81.0
& 2.14
& 59.8
\\

\textbf{SFT + GRPO + Reflection + Memory}
& \checkmark
& \checkmark
& \checkmark
& \checkmark
& \textbf{94.9}
& \textbf{90.9}
& \textbf{87.4}
& \textbf{89.6}
& \textbf{88.5}
& \textbf{81.3}
& \textbf{2.21}
& \textbf{60.4}
\\

\bottomrule
\end{tabular}
\end{table*}

\subsubsection{Evaluation Metrics}
\noindent\textbf{\emph{Narrative Quality (NQ)}} measures whether the video forms
a complete, coherent, and engaging advertising narrative. It averages
narrative completeness (logical organization, information density,
opening/ending) and attractiveness (attention capture, rhythm,
memorability), each scored in $[0,50]$.

\noindent\textbf{\emph{Selling-Point Coverage (SPC)}} evaluates how completely
the video communicates the core selling points of the target product. We
first extract candidate selling points from the product detail page and
retain only those that can be visually demonstrated or explicitly
conveyed in a short advertising video. For each retained selling point,
the evaluator examines the visual frames, narration, subtitles, and
on-screen text and assigns a matching score $m_k\in[0,1]$. A selling
point is regarded as covered when its matching score exceeds a fixed
threshold $\tau$. Given $K$ retained selling points, SPC is calculated
as
$
\mathrm{SPC}
=
\frac{1}{K}
\sum_{k=1}^{K}
\mathbb{I}(m_k\geq\tau)
\times 100.
$

\noindent\textbf{\emph{Visual--Script Consistency (VSC)}} evaluates the semantic
consistency between visuals and corresponding narration/subtitles.
ASR-aligned transcript segments are compared with co-temporal frames by
the multi-modal evaluator.

\noindent\textbf{\emph{Material Quality Assessment (MQA)}} evaluates the
clarity, relevance, and diversity of the visual materials, penalizing
blur, noise, exposure issues, irrelevant content, and cross-shot
redundancy.

\noindent\textbf{\emph{Music Similarity Score (MSS)}} evaluates whether the
background music is compatible with the product category, visual content,
and editing rhythm. We infer the desired music style from frames and
subtitles, separate the background music from speech, and measure
compatibility via CLAP embedding cosine similarity:
\begin{equation}
s_{\mathrm{MSS}}
=
\cos\left(
f_{\mathrm{CLAP}}(a_{\mathrm{bgm}}),
f_{\mathrm{CLAP}}(t_{\mathrm{style}})
\right).
\end{equation}

\noindent\textbf{\emph{Trust Building (TB)}} evaluates whether the video makes
the product appear reliable and worth purchasing, combining scene
immersion (realistic usage scenarios) and trust endorsement
(certifications, reviews, demonstrations), each scored in $[0,50]$.

\subsubsection{Quantitative Comparisons}
We evaluate all methods on CutAgent-Eval, as summarized in Table~\ref{tab:main_comparison}. All methods receive identical multi-modal product inputs and clip pools; gene-generation baselines share our VideoAgent for downstream production, while end-to-end baselines use their own editors. The generated videos are comprehensively assessed across three paradigms: a six-dimensional automated quality protocol (scored on a 0--100 scale using Qwen3.5-397B-A17B~\cite{bai2025qwen3}; details in Sec.~\ref{app:metrics}), real-world online CTR, and a forced-choice human preference study. In
particular, compared with AutoCut, which maps product inputs directly to
video-production decisions without an explicit creative-gene planning
stage, CutAgent attains consistently better automated quality, the
highest scores in five of the six automated quality dimensions, the
highest human preference (60.4\%), and the best online CTR
(\textbf{2.21\%}).

\subsubsection{Qualitative Comparison}
Fig.~\ref{fig:qualitative_comparison} in Sec.~\ref{app:qualitative} compares the two systems shot by shot on two
representative products. AutoCut opens with near-duplicate shots and
flattens into a selling-point list, whereas CutAgent orders attention
capture, product recognition, feature demonstration, and trust building,
covering complementary product evidence rather than redundant close-ups,
and raises the observed CTR from 1.83\% to 2.89\% and from 1.27\% to
2.81\%, suggesting that the gain stems from narrative ordering rather
than from shot-level aesthetics.

\subsubsection{{Online CTR}} 
To evaluate real-world advertising effectiveness, we ran a one-week online
experiment on a large e-commerce platform covering 43 categories and 430
products (10 per category), with over 10 million ad impressions. All
methods share same products and randomly allocated traffic; the
LightGBM predictor of Sec.~\ref{sec:ctr_reward} serves only as a
training-time reward.

\subsubsection{Human Evaluation}
We conduct a forced-choice A/B study with 200 products sampled from
CutAgent-Eval and 30 participants, collecting 1{,}600 pairwise judgments
(200 products $\times$ 8 methods). Each baseline is compared with
CutAgent on the same products, and CutAgent's overall preference score
averages its forced-choice win rate over the eight methods.
Participants choose without ties, considering all six dimensions and
overall appeal (details in Sec.~\ref{app:metrics}).

\subsubsection{Ablation Study}
Table~\ref{tab:cutagent_ablation} evaluates SFT, CTR-guided GRPO,
reflection, and long-term memory; the Base variant uses
Qwen3-VL-8B-Instruct without task-specific optimization.
SFT substantially improves all metrics: NQ/SPC rise from 86.8/79.6 to
92.0/87.3, TB from 66.8 to 76.2, online CTR from 1.12\% to 1.76\%, and
human preference from 37.5\% to 54.2\%. Reflection mainly improves
narrative and post-production quality: SFT+Reflection reaches NQ 94.6,
SPC 90.1, MQA 88.9, TB 81.5, preference 59.1\%, and CTR 1.89\%. GRPO
instead improves advertising effectiveness and cross-modal alignment:
SFT+GRPO reaches CTR 2.05\%, with VSC/MSS increasing from 85.4/87.2 to
87.3/88.7.
Combining both yields NQ 94.7, SPC 90.7, MQA 89.7, preference 59.8\%,
and CTR 2.14\%. Long-term memory adds smaller but consistent gains: NQ,
SPC, VSC, MSS, and TB improve by 0.2, 0.2, 0.2, 0.2, and 0.3 points,
raising CTR and preference to 2.21\% and 60.4\%.

\section{Conclusion}

We presented CutAgent, a dual-agent framework that formulates e-commerce
advertising video creation as a Plan--Execute--Self-Evolve process.
GeneAgent maps multi-modal product inputs---textual attributes together
with product images---to structured creative genes through high-CTR
supervised fine-tuning and CTR-guided GRPO on CutAgent-70K, whose genes
are recovered by reverse extraction from historical ads. VideoAgent
converts these genes into complete advertising videos under
category-then-gene constraints from a long-term memory $\mathcal{E}$,
and independently refines production decisions via reflection, writing
verified corrections back to $\mathcal{E}$. Experiments on CutAgent-Eval
show that CutAgent consistently outperforms representative open-source
and proprietary baselines in video quality, online CTR, and human
preference, achieving the highest online CTR of 2.21\%.

{
    \small
    \bibliographystyle{ieeenat_fullname}
    \bibliography{main}
}
\clearpage
\appendix

\section{GeneAgent Training Details}
\label{app:impl}
\subsection{Narrative Unit Registry}
\label{app:nu_registry}

Table~\ref{tab:nu_registry} presents the complete narrative-unit
vocabulary used by CutAgent. The vocabulary contains 27 narrative
units organized into six functional families: attention capture,
product introduction, functional demonstration, trust building,
promotional conversion, and closing.

\begin{table*}[p]
\centering
\caption{Complete narrative-unit registry used by CutAgent.}
\label{tab:nu_registry}
\footnotesize
\setlength{\tabcolsep}{4pt}
\renewcommand{\arraystretch}{1.00}
\begin{tabular}{
    >{\raggedright\arraybackslash}p{1.1cm}
    >{\raggedright\arraybackslash}p{3.45cm}
    >{\raggedright\arraybackslash}p{6.65cm}
    >{\raggedright\arraybackslash}p{4.9cm}
}
\toprule
Code & Narrative Unit & Description & Typical Scene \\
\midrule
\multicolumn{4}{l}{\textit{Attention Capture}}\\
NU-L1 & Price Shock & Highlights an unexpectedly attractive price to create immediate surprise. & Large price caption, such as ``Only \$9.9.'' \\
NU-B3 & Trend-Based Opening & Introduces the product through a current trend, popular topic, or Internet meme. & Trending scenarios or popular online expressions. \\
NU-P0 & Visual Impact & Captures attention through striking product effects, transformations, or contrasts. & Dynamic product presentation or before--after comparison. \\
NU-O4 & Suspense Preview & Presents the final result first and then reveals how it is achieved. & An impressive result followed by a suspenseful question. \\
NU-D3 & Pain-Point Scenario & Presents the inconvenience experienced before using the product. & A realistic scene in which the user encounters a problem. \\
\addlinespace
\multicolumn{4}{l}{\textit{Product Introduction}}\\
NU-Q7 & Full Product View & Presents the complete appearance and overall design of the product. & Front, panoramic, or multi-angle product views. \\
NU-C1 & Detail Close-Up & Highlights the material, craftsmanship, texture, or local product details. & Macro or close-up shots of product details. \\
NU-T8 & Unboxing & Shows the unpacking and initial product-reveal process. & Opening the package and removing the product. \\
NU-D7 & Package \& Accessories & Presents the packaging, included accessories, and complimentary items. & Flat-lay display of the package and accessories. \\
NU-G8 & Specifications & Displays product parameters, ingredients, dimensions, or specifications. & Close-up of a specification label or ingredient list. \\
NU-U9 & Product Variants & Presents multiple colors, styles, sizes, or product variants. & Multiple SKUs displayed side by side or sequentially. \\
\addlinespace
\multicolumn{4}{l}{\textit{Functional Demonstration}}\\
NU-H9 & Core Function Demo & Demonstrates one primary function of the product. & Operating the product to show its core function. \\
NU-N5 & Usage Tutorial & Explains how to use the product through sequential operating steps. & Step-by-step product-use demonstration. \\
NU-R9 & Effect Demonstration & Presents the immediate outcome or effect produced by the product. & Before--after comparison or direct result display. \\
NU-X2 & Usage Scenario & Demonstrates product use in a realistic application scenario. & Using the product in a kitchen, bedroom, or outdoor environment. \\
NU-Z2 & Multi-Scenario Demo & Shows that the product can be applied across multiple usage scenarios. & Switching between different product-use environments. \\
NU-S6 & A/B Comparison & Compares the product with a competing product or conventional solution. & Split-screen or sequential comparison. \\
\addlinespace
\multicolumn{4}{l}{\textit{Trust Building}}\\
NU-A7 & Influencer Endorsement & Builds credibility through a recommendation from an influencer or expert. & An influencer appears on screen and recommends the product. \\
NU-V4 & User Testimonial & Uses customer feedback or testimonials to establish product credibility. & Customer-review screenshots or on-camera testimonials. \\
NU-I6 & Authoritative Certification & Presents certificates, test reports, or authoritative endorsements. & Close-up of a certificate or inspection report. \\
NU-F5 & Technology Explanation & Explains the product's ingredients, mechanism, or underlying technology. & Ingredient lists accompanied by explanatory graphics. \\
NU-Y3 & Brand Story \& Provenance & Introduces the brand history, manufacturing process, or product origin. & Factory, place-of-origin, or brand-history footage. \\
\addlinespace
\multicolumn{4}{l}{\textit{Promotional Conversion}}\\
NU-W6 & Price/Discount Reveal & Reveals the product price, discount, or promotional offer. & Animated price label or spoken price announcement. \\
NU-J4 & Limited-Time Promotion & Creates urgency through a limited-time or limited-stock offer. & Countdown timer or limited-stock label. \\
NU-K3 & Gift or Bundle Value & Highlights complimentary items, bundles, or package value. & Display of free gifts or a complete product bundle. \\
NU-M2 & Purchase Guidance & Explicitly directs the viewer toward the purchase action. & Arrow animation accompanied by ``Click the link.'' \\
\addlinespace
\multicolumn{4}{l}{\textit{Closing}}\\
NU-E0 & Branded End Card & Concludes the video with brand identity and a consistent visual ending. & Brand logo animation accompanied by a slogan. \\
\bottomrule
\end{tabular}
\end{table*}

\subsection{Hyperparameters}

We adopt Qwen3-VL-8B-Instruct as the backbone of GeneAgent. During both
SFT and GRPO, each input is the multimodal product representation $P_i$:
textual product attributes (category, SKU title, price, selling points,
and review summary) together with the main and detail product images.
The gene-generation instruction follows the creative-gene prompt in
Fig.~\ref{fig:prompts}(A).

We freeze the visual encoder and the vision--language alignment module,
and apply LoRA to all linear layers of the language model. The LoRA rank,
scaling factor, and dropout rate are set to 16, 32, and 0.05,
respectively. GeneAgent is first supervised fine-tuned on the top 20\%
CTR samples of CutAgent-70K for three epochs using
AdamW~\cite{loshchilov2017decoupled} with an initial learning rate of
$5\times10^{-5}$, a weight decay of 0.01, and a cosine learning-rate
schedule with a warm-up ratio of 0.05. The effective batch size is 32,
the maximum sequence length is 4{,}096, and the maximum gradient norm is
1.0. We use BF16 precision~\cite{kalamkar2019study},
FlashAttention~\cite{dao2022flashattention}, and gradient checkpointing.
The training loss is computed only over the target creative-gene tokens.

Starting from the SFT checkpoint, we further optimize GeneAgent with
CTR-guided GRPO on the full CutAgent-70K for one epoch. For each product
we sample $K{=}8$ candidate genes with temperature $1.0$ and
top-$p{=}0.9$, and score them with the composite reward. We use AdamW
with learning rate $1\times10^{-6}$, effective batch size 32 (1 product
$\times$ 8 candidates per step when accumulated), clipping coefficient
$\epsilon{=}0.2$, KL coefficient $\beta{=}0.04$, and advantage
stabilizer $\delta{=}10^{-6}$.

\subsection{CTR-Level Reward Model Hyperparameters}
\label{app:ctr_reward_hparams}

We use the multiclass objective with learning rate 0.05, 31 leaves,
maximum depth 8, 500 boosting rounds, and feature/sample subsampling
ratios of 0.8. The minimum number of samples per leaf is 20, and
inverse-frequency class weights alleviate imbalance.

We further evaluate the LightGBM-based CTR-level reward model used
\emph{only during training} in Table~\ref{tab:ctr_reward_model}. The
model achieves an accuracy of 87.6\% on the CutAgent-70K training set
and 76.9\% on an independent held-out split from CutAgent-Eval product
attributes and genes. Compared with the 60\% majority-class baseline
from the imbalanced CTR-level distribution, the reward model obtains a
substantial improvement on unseen samples, indicating that it captures
the relationship between product attributes, creative genes, and
historical CTR levels for GRPO. This surrogate is never used to report
the online CTR results in our main comparison or ablation tables.
\begin{table}[h]
\centering
\small
\caption{
Accuracy of the training-time LightGBM CTR-level reward model on
CutAgent-70K and a held-out gene-level test split.
}
\label{tab:ctr_reward_model}
\setlength{\tabcolsep}{10pt}
\renewcommand{\arraystretch}{1.10}
\begin{tabular}{lcc}
\toprule
Dataset & Split & Accuracy (\%)$\uparrow$ \\
\midrule
CutAgent-70K  & Train & 87.6 \\
CutAgent-Eval & Test  & 76.9 \\
\bottomrule
\end{tabular}
\end{table}

\subsection{CTR-Level Discretization}
\label{app:grpo_details}

Raw CTR distributions vary substantially across categories. We rank
historical advertisements separately within each category and divide them
into Low-, Mid-, and High-CTR levels:
\begin{equation}
z_i
=
\begin{cases}
\mathrm{low},
&
y_i < Q_{0.30}^{(c_i)},\\
\mathrm{mid},
&
Q_{0.30}^{(c_i)}
\leq y_i < Q_{0.90}^{(c_i)},\\
\mathrm{high},
&
y_i \geq Q_{0.90}^{(c_i)},
\end{cases}
\end{equation}
where $Q_{\alpha}^{(c_i)}$ is the $\alpha$-quantile of CTR within
category $c_i$. This produces approximate proportions of 30\%, 60\%,
and 10\%.

\subsection{Feature Construction and Reward Model}

For each historical product--gene pair, we construct
\begin{equation}
\mathbf{x}_i
=
\phi(P_i,G_i)
\in\mathbb{R}^{d},
\qquad
d\approx250.
\end{equation}
The features include the narrative-unit sequence, keywords from all gene
rules, TF--IDF representations of textual fields, and product attributes
such as category and price. A LightGBM~\cite{ke2017lightgbm} classifier
estimates
\begin{equation}
q_{\psi}
\left(
z\mid P_i,G_i
\right)
=
q_{\psi}
\left(
z\mid\phi(P_i,G_i)
\right),
\end{equation}
using the multiclass objective
\begin{equation}
\mathcal{L}_{\mathrm{RM}}
=
-\frac{1}{N}
\sum_{i=1}^{N}
\log
q_{\psi}
\left(
z_i\mid\phi(P_i,G_i)
\right).
\end{equation}
The trained classifier is frozen throughout GRPO.

\subsection{Reward Components}
\label{app:reward_components}

For each product input $P_i$ (textual attributes and product images),
the old policy samples $K$ candidate creative genes:
\begin{equation}
\widehat{G}_i^{(k)}
\sim
\pi_{\theta_{\mathrm{old}}}
\left(
G\mid P_i
\right),
\qquad
k=1,\ldots,K.
\end{equation}
During training, the frozen CTR-level predictor assigns each candidate a
CTR-level class:
\begin{equation}
\widehat{z}_i^{(k)}
=
\arg\max_{z\in\mathcal{Z}}
q_{\psi}
\left(
z\mid
\phi(P_i,\widehat{G}_i^{(k)})
\right),
\end{equation}
where
$\mathcal{Z}=\{\mathrm{low},\mathrm{mid},\mathrm{high}\}$.

The three levels receive rewards of 0, 2, and 5:
\begin{equation}
R_{\mathrm{ctr},i}^{(k)}
=
\begin{cases}
0, & \widehat{z}_i^{(k)}=\mathrm{low},\\
2, & \widehat{z}_i^{(k)}=\mathrm{mid},\\
5, & \widehat{z}_i^{(k)}=\mathrm{high}.
\end{cases}
\end{equation}
We normalize this reward as
\begin{equation}
\widetilde{R}_{\mathrm{ctr},i}^{(k)}
=
\frac{R_{\mathrm{ctr},i}^{(k)}}{5}.
\end{equation}

A format-compliance reward
$R_{\mathrm{fmt},i}^{(k)}\in[0,1]$ evaluates field completeness,
serialization correctness, and narrative-unit consistency. A gene-diversity
reward discourages repetitive candidates:
\begin{equation}
R_{\mathrm{div},i}^{(k)}
=
1-
\frac{1}{K-1}
\sum_{\substack{j=1\\j\neq k}}^{K}
\mathrm{sim}
\left(
\widehat{G}_i^{(k)},
\widehat{G}_i^{(j)}
\right),
\end{equation}
where $\mathrm{sim}(\cdot,\cdot)$ is the mean Jaccard similarity over
narrative-unit sets and discrete production-rule fields.

\subsection{Group-Relative Advantage and Policy Optimization}
\label{app:grpo_adv}

For candidates generated from the same product, the group-relative
advantage is
\begin{equation}
\widehat{A}_i^{(k)}
=
\frac{
R_i^{(k)}-\mu_i
}{
\sigma_i+\delta
},
\end{equation}
where
\begin{equation}
\mu_i
=
\frac{1}{K}\sum_{k=1}^{K}R_i^{(k)},
\qquad
\sigma_i
=
\sqrt{
\frac{1}{K}
\sum_{k=1}^{K}
\left(
R_i^{(k)}-\mu_i
\right)^2
}.
\end{equation}

Let $\widehat{g}_{i,k,t}$ denote the $t$-th token of
$\widehat{G}_i^{(k)}$. The token-level policy ratio is
\begin{equation}
\rho_{i,k,t}(\theta)
=
\frac{
\pi_{\theta}
\left(
\widehat{g}_{i,k,t}
\mid
P_i,\widehat{G}_{i,<t}^{(k)}
\right)
}{
\pi_{\theta_{\mathrm{old}}}
\left(
\widehat{g}_{i,k,t}
\mid
P_i,\widehat{G}_{i,<t}^{(k)}
\right)
}.
\end{equation}

The KL term is computed against the frozen SFT reference policy:
\begin{equation}
D_{\mathrm{KL}}^{i,k,t}
=
D_{\mathrm{KL}}
\left(
\pi_{\theta}
\left(
\cdot\mid
P_i,\widehat{G}_{i,<t}^{(k)}
\right)
\middle\|
\pi_{\mathrm{ref}}
\left(
\cdot\mid
P_i,\widehat{G}_{i,<t}^{(k)}
\right)
\right),
\end{equation}
where $\pi_{\theta_{\mathrm{old}}}$ generates the current candidate
group, while $\pi_{\mathrm{ref}}$ denotes the fixed SFT model.
$\delta{=}10^{-6}$ stabilizes the group-relative advantage.

\section{VideoAgent Production and Reflection Details}
\label{app:videoagent}

Guided by the creative gene, VideoAgent retrieves experiences from the
long-term memory $\mathcal{E}$ by first selecting the product's category
shard among the 43 categories, then ranking gene contexts inside that
shard with BGE-M3~\cite{chen2024bge} and injecting the top-$K$ matches
as constraints for script expansion. The VideoAgent LLM prompts in
Fig.~\ref{fig:prompts}(B) are executed by
Qwen3-235B-A22B~\cite{yang2025qwen3}: B1 expands the gene into a
shot-by-shot script, and B2 adapts narration and subtitles for TTS.
Video-clip retrieval uses Qwen3-VL-Embedding-8B~\cite{li2026qwen3} over
representative frames of candidate clips. TTS is performed with
CosyVoice~2~\cite{du2024cosyvoice}, and word-level timestamps construct
subtitle events. BGM retrieval combines
Qwen3-Embedding-8B~\cite{li2026qwen3} semantic matching with acoustic
features extracted by librosa~\cite{mcfee2015librosa}, including tempo,
energy, and spectral statistics. The final timeline is rendered with
ffmpeg~\cite{tomar2006converting}. After rendering, VideoAgent runs the
independent reflection loop
(evaluation--diagnosis--revision--verification) for at most three rounds
and writes verified failure--correction pairs back to $\mathcal{E}$ for
subsequent category-then-gene retrieval.

\subsection{Video-Clip Retrieval}

For each shot, candidate clips are retrieved from the product-specific
cleaned material pool $\mathcal{M}_i$ by encoding representative frames
with a vision--language embedding model~\cite{li2026qwen3} and ranking by similarity to the
shot description, while encouraging cross-shot visual diversity. Because
the shot description is planned from $P_i$, including product images,
retrieval favors clips that match both the narrative intent and the
product appearance.

\subsection{TTS and Subtitle Generation}

Following the B2 prompt in Fig.~\ref{fig:prompts}, VideoAgent adapts the
shot scripts and voice-over rule into TTS-ready narration and subtitle
styles, then synthesizes speech; word-level timestamps construct
subtitle events. The script is shortened when its duration is
inconsistent with the selected clip. Subtitle images are burned in at
render time; unifying subtitle synthesis with visual text generation
and editing, as in InnoText~\cite{liu2026innotext}, is a direct route
to more expressive on-screen typography
(Sec.~\ref{app:gen_discussion}).

\subsection{BGM and Sound-Effect Matching}

Background music is retrieved by the music rule using semantic
similarity and acoustic cues (tempo, energy, spectrum, emotion). Sound
effects follow shot actions and the sound-effect rule.

\subsection{Timeline Rendering}

Selected clips, narration, subtitles, BGM, sound effects, and
transitions are aligned on a timeline and rendered into the final
video. Retrieved clips could in principle be replaced or supplemented
by synthesized shots; doing so without subject drift or abrupt
transitions is exactly the setting studied by
FancyVideo~\cite{feng2025fancyvideo} and
RelaCtrl~\cite{cao2026relactrl} (Sec.~\ref{app:gen_discussion}).

\subsection{Reflection Loop}
\label{app:reflection}

VideoAgent performs a four-stage reflection loop on the rendered video.

\noindent\textbf{Evaluation.}
The rendered video is evaluated from six fine-grained dimensions:
narrative quality (NQ), selling-point coverage (SPC), visual--script
consistency (VSC), material quality assessment (MQA), music similarity
score (MSS), and trust building (TB). In addition to scalar scores, the
evaluator produces temporally grounded evidence for individual shots,
texts, audio segments, and editing elements.

\noindent\textbf{Diagnosis.}
For each dimension that fails to satisfy its quality requirement,
VideoAgent analyzes the evaluation evidence and localizes the defect to
a specific temporal interval, shot instruction, or production module,
such as a repeated clip, omitted selling point, mismatched narration,
low-quality material, unsuitable music, or abrupt transition.

\noindent\textbf{Revision.}
Based on the diagnosis, VideoAgent generates targeted correction
instructions and updates the corresponding shot script. Rather than
regenerating the entire video, it preferentially modifies only the
affected components---for example, replacing a clip, rewriting a
voice-over segment, adjusting shot duration, or rematching the
background music---and re-executes the relevant production modules.

\noindent\textbf{Verification.}
The revised video is evaluated again under the same protocol. If one or
more quality requirements remain unsatisfied, the updated evidence is
passed to the next diagnosis--revision cycle. Otherwise, the corrected
video is accepted, and any verified failure--correction pair is written
into the long-term memory for reuse on subsequent genes.

\section{Dataset Construction Details}
\label{app:dataset}

\begin{figure}[t]
  \centering
  \includegraphics[width=1\linewidth]{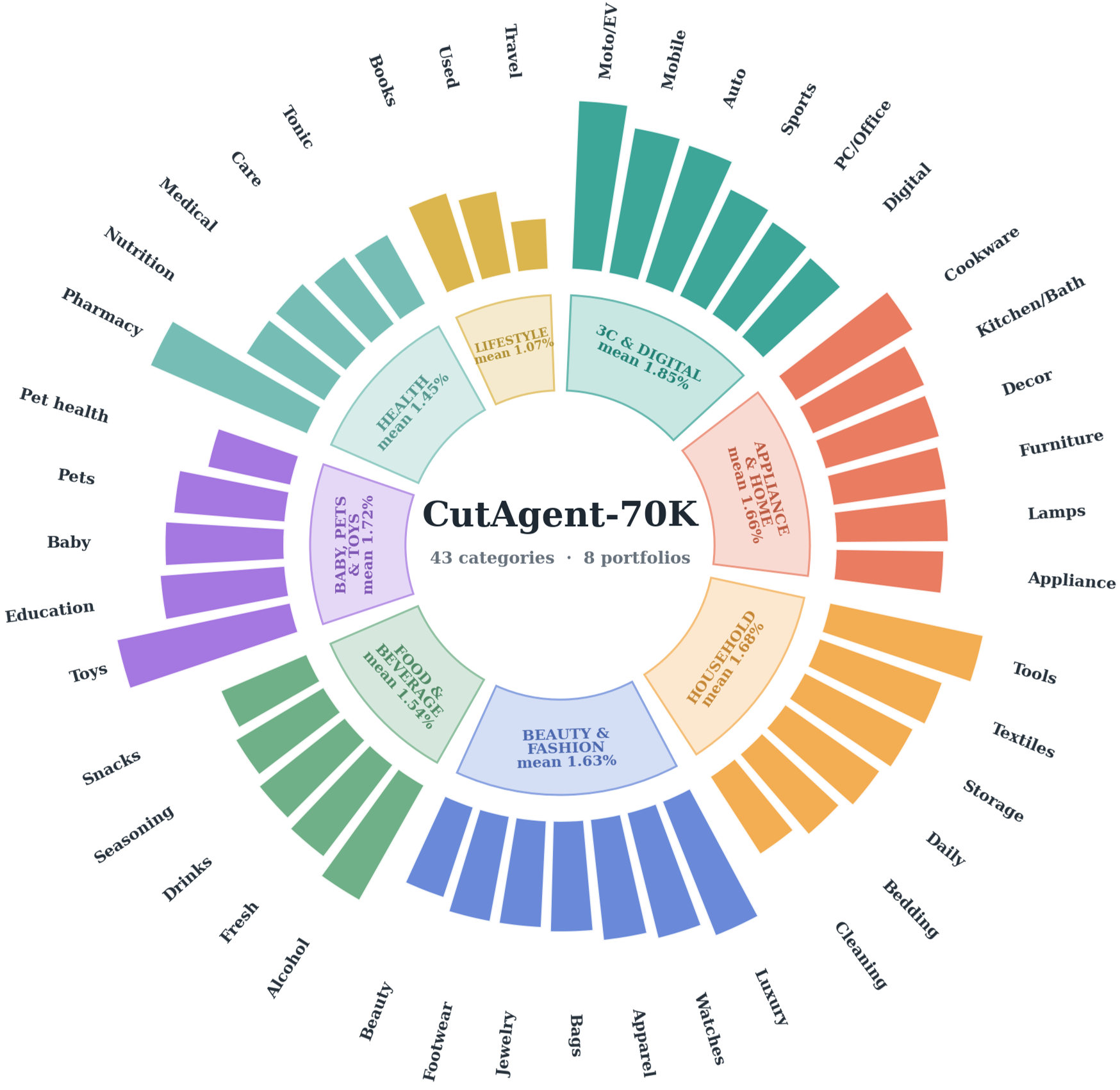}
  \caption{
  Category distribution of CutAgent-70K across 43 categories and 8
  portfolios. Outer bars indicate relative category mass; inner labels
  report portfolio-level mean share.
  }
  \label{fig:dataset_sta}
\end{figure}

\noindent\textbf{Category Distribution.}
CutAgent-70K covers 43 product categories organized into 8 portfolios,
as shown in Fig.~\ref{fig:dataset_sta}: 3C \& Digital, Appliance \&
Home, Household, Beauty \& Fashion, Food \& Beverage, {Baby, Pets \&
Toys}, Health, and Lifestyle. The outer radial bars indicate relative
category mass within each portfolio, while the inner ring reports
portfolio-level mean share, showing broad but non-uniform coverage across
major e-commerce verticals.

\noindent\textbf{Material Cleaning Pipeline.}
To support VideoAgent retrieval, historical advertising videos are
processed by a lightweight material-cleaning pipeline into a reusable
clip pool:
\begin{enumerate}
\item \emph{Scene detection}: ffmpeg splits each video into shots based
on scene changes.
\item \emph{Quality filtering}: clips containing on-camera humans or
lasting under 2\,s are removed.
\item \emph{Subtitle/watermark removal}: overlaid subtitles and
watermarks are detected and blurred while preserving on-product printed
text.
\item \emph{Deduplication}: near-duplicate clips are eliminated by
perceptual hashing.
\end{enumerate}
After cleaning, each product typically yields 90--120 usable clips,
denoted by $\mathcal{M}_i$.

\noindent\textbf{Reverse Gene Extraction.}
Historical creative genes are recovered from the original advertising
videos by the reverse-extraction pipeline in Fig.~\ref{fig:prompts}(C),
which produces the supervision targets $G_i$ of CutAgent-70K (distinct
from the runtime GeneAgent prompt in Fig.~\ref{fig:prompts}(A)).

\emph{Voice attribute analysis} (C1) and \emph{BGM style analysis} (C2)
use Qwen3.5-Omni-Plus-Realtime on separated audio to obtain vocal
attributes (e.g., gender, pitch, speed, expression) and a compact BGM
style summary; these audio attributes later ground the voice-over rule
$r_{v,i}$ and music rule $r_{m,i}$.

\emph{Shot-level multimodal analysis} (C3) uses
Doubao-Seed-2.0-pro~\cite{seed2026seed2} (vision--language) on each shot
and outputs visual content, oral content, and narrative purpose, with
voice-over as the primary cue and visuals as auxiliary evidence.

\emph{Creative gene reconstruction} (C4) uses Doubao-Seed-2.0-pro (text)
with the shot-level analysis, the narrative-unit registry, and the audio
attributes to restore a full creative gene in the gene schema, including
the ordered narrative units
$S_i=[s_{i,1},\ldots,s_{i,n_i}]$ ($s_{i,j}\in\mathcal{NU}$) and the
production rules $r_{v,i}$, $r_{t,i}$, $r_{m,i}$, $r_{s,i}$, $r_{f,i}$.
Assembling these fields gives
$G_i=(S_i,r_{v,i},r_{t,i},r_{m,i},r_{s,i},r_{f,i})$.

\begin{figure*}[t]
  \centering
  \includegraphics[width=0.8\linewidth]{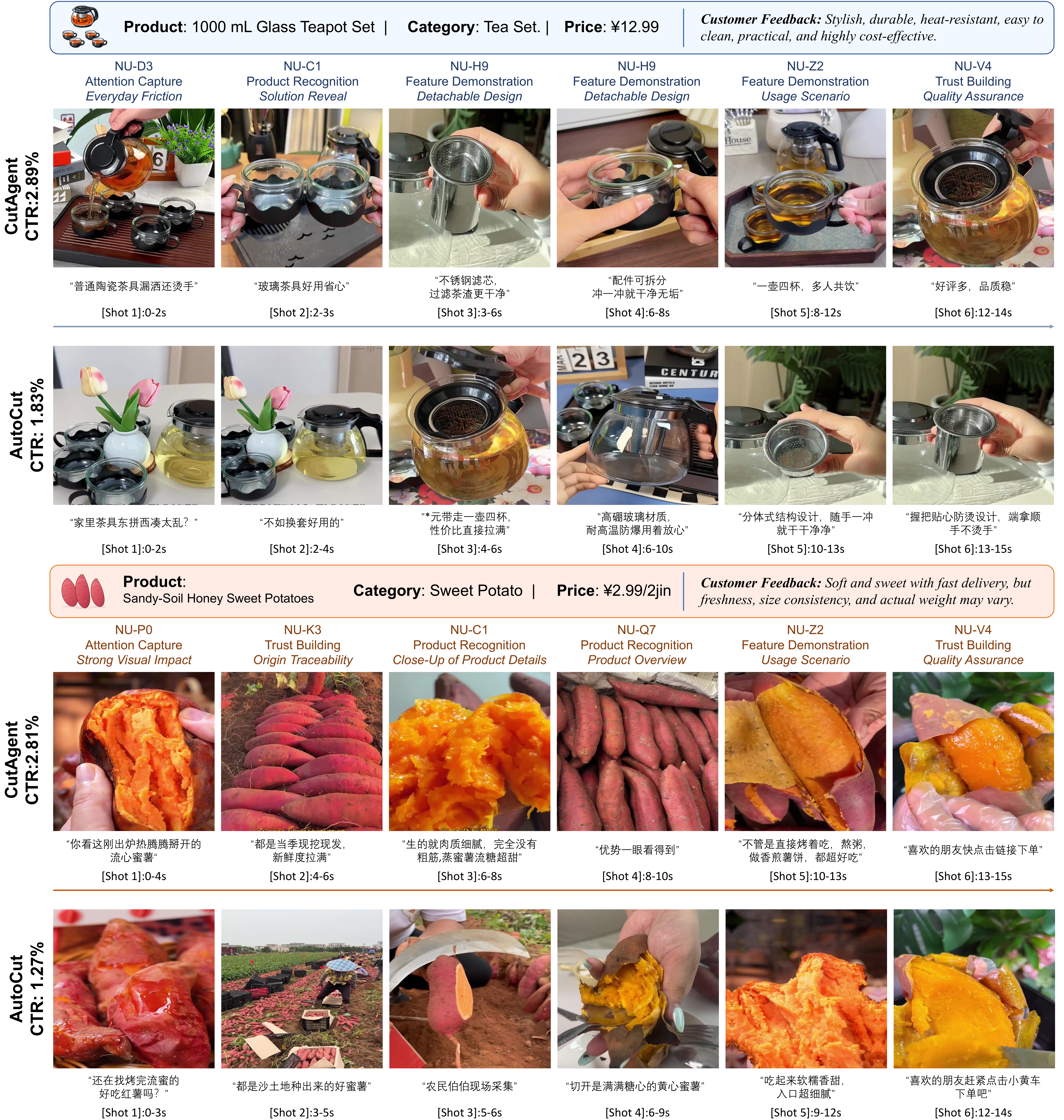}
  \caption{
  Qualitative comparison between AutoCut and CutAgent on a glass teapot
  set and sandy-soil honey sweet potatoes.
  }
  \label{fig:qualitative_comparison}
\end{figure*}

\section{Evaluation Metric Definitions}
\label{app:metrics}

\subsection{Narrative Quality (NQ)}

NQ measures whether the final video forms a complete, coherent, and
engaging advertising narrative rather than a simple concatenation of
product materials. Narrative completeness evaluates the clarity of the
narrative progression, logical organization of product information,
information density, and the naturalness of the opening and ending.
Attractiveness evaluates whether the opening captures attention, whether
the intermediate content maintains an appropriate rhythm, and whether the
video contains memorable or appealing elements. Each aspect is scored
from $0$ to $50$, and NQ is computed as
\begin{equation}
\mathrm{NQ}
=
S_{\mathrm{completeness}}
+
S_{\mathrm{attractiveness}},
\end{equation}
where
$S_{\mathrm{completeness}},S_{\mathrm{attractiveness}}\in[0,50]$.

\subsection{Visual--Script Consistency (VSC)}

Inspired by the Visual--Script Correlation metric in
AutoCut~\cite{zhou2026autocut}, we adapt it to reference-free evaluation
of complete advertising videos. We apply automatic speech recognition to
obtain timestamped narration transcripts, align each text segment with
the visual frames in the corresponding temporal interval, and ask the
multimodal evaluator to assess whether the visual content conveys the
expressed meaning and whether subtitles reinforce the visual focus.
These aspects are jointly considered to produce a VSC score between $0$
and $100$. Because spatial reasoning remains a weak point of current
MLLMs~\cite{lu2026space}, VSC is best read as a relative signal between
methods rather than an absolute measurement, and we therefore report it
together with the forced-choice human study of
Sec.~\ref{app:human}.

\subsection{Material Quality Assessment (MQA)}

MQA evaluates the quality, relevance, and diversity of the visual
materials used in the final video, including image clarity, exposure
quality, positive product presentation, relevance to the target product
and its selling points, and unnecessary repetition across clips.
Representative frames from each detected shot are examined for blur,
noise, over-/underexposure, irrelevant content, and redundant materials.
The evaluator then assigns an overall score between $0$ and $100$.

\subsection{Music Similarity Score (MSS)}

Following AutoCut~\cite{zhou2026autocut}, we adapt it to reference-free
music--video evaluation. We use a multimodal evaluator to infer the
desired music style from frames and subtitles, separate the background
music from speech, and encode both the extracted music and the style
description with CLAP~\cite{wu2023large}. The compatibility score is mapped to the
$0$--$100$ range with the same rule for all methods.

\begin{figure*}[t]
  \centering
  \includegraphics[width=0.7\linewidth,keepaspectratio]{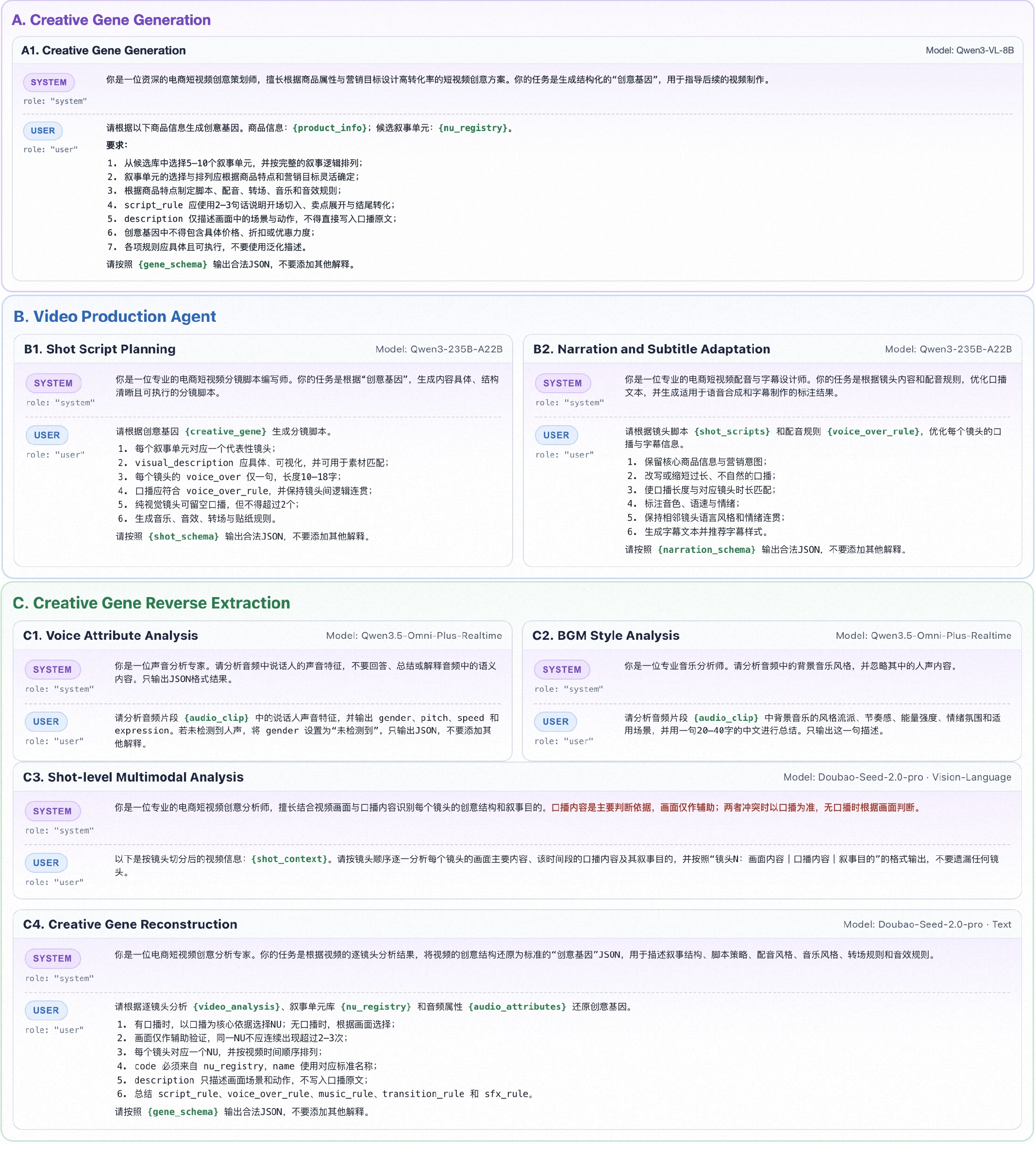}
  \caption{
  Prompt templates used in CutAgent.
  (A)~Creative-gene generation prompts used by GeneAgent at training and
  inference.
  (B)~VideoAgent prompts for shot-script planning (B1) and
  narration/subtitle adaptation (B2).
  (C)~Reverse-extraction prompts that recover creative genes from
  historical advertising videos for CutAgent-70K supervision: voice
  attribute analysis (C1), BGM style analysis (C2), shot-level multimodal
  analysis (C3), and creative-gene reconstruction (C4).
  }
  \label{fig:prompts}
\end{figure*}

\subsection{Trust Building (TB)}

TB evaluates whether the advertising video makes the product appear
reliable and worth purchasing. Scene immersion measures whether the video
presents realistic usage scenarios and processes that viewers can imagine
themselves in. Trust endorsement measures supporting evidence such as
certifications, reviews, sales statistics, or real-person demonstrations.
Each aspect is scored from $0$ to $50$, and
\begin{equation}
\mathrm{TB}
=
S_{\mathrm{scene}}
+
S_{\mathrm{endorsement}},
\end{equation}
where $S_{\mathrm{scene}},S_{\mathrm{endorsement}}\in[0,50]$.

\subsection{Online CTR}

For method $m$, online CTR is computed from the observed clicks and
impressions:
\begin{equation}
\mathrm{CTR}_{m}
=
\frac{
\sum_{i=1}^{N} C_i^{m}
}{
\sum_{i=1}^{N} E_i^{m}
}
\times 100\%,
\end{equation}
where $C_i^{m}$ and $E_i^{m}$ denote the numbers of clicks and
impressions received by the video generated for product $i$,
respectively.

\subsection{Human Evaluation}
\label{app:human}

For a competing method $m$,
\begin{equation}
\mathrm{Pref.}(m)
=
\frac{N_m}
{N_m+N_{\mathrm{CutAgent}}}
\times 100\%,
\end{equation}
where $N_m$ and $N_{\mathrm{CutAgent}}$ denote the numbers of votes
received by method $m$ and CutAgent, respectively. Since each comparison
uses a forced-choice protocol, the two preference rates sum to $100\%$.
When all competing methods are evaluated using the same number of
pairwise judgments, the overall preference score of CutAgent is
\begin{equation}
\mathrm{Pref.}(\mathrm{CutAgent})
=
\frac{1}{M}
\sum_{m=1}^{M}
\left(
100\%-\mathrm{Pref.}(m)
\right),
\end{equation}
where $M$ denotes the number of competing methods.

\begin{figure*}[t]
  \centering
  \includegraphics[width=0.95\linewidth,keepaspectratio]{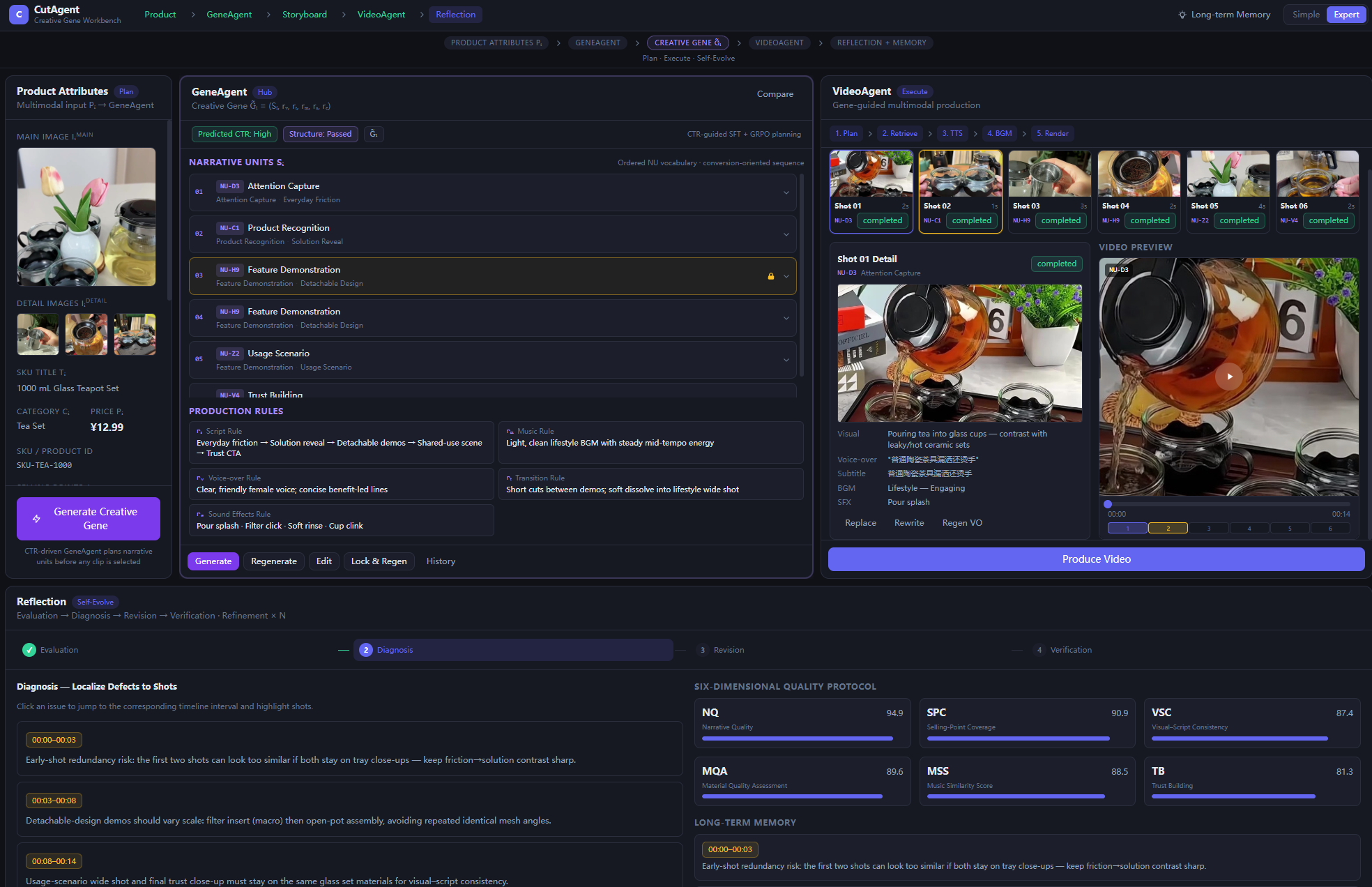}
  \caption{
  CutAgent system interface for e-commerce advertising video creation.
  From left to right, the workbench exposes multimodal product attributes
  $P_i$ (main/detail images and textual metadata), the GeneAgent hub that
  edits the structured creative gene $\widetilde{G}_i$ (narrative units
  and production rules), and the VideoAgent panel for memory-constrained
  shot planning, clip retrieval, TTS, BGM matching, and rendering.
  The bottom Reflection panel visualizes the six-dimensional evaluation
  (NQ, SPC, VSC, MQA, MSS, TB) together with temporally grounded
  diagnosis and revision, closing the Plan--Execute--Self-Evolve loop.
  }
  \label{fig:system}
\end{figure*}

\section{Qualitative Analysis}
\label{app:qualitative}

We provide a detailed shot-by-shot comparison between AutoCut and
CutAgent on two representative products from CutAgent-Eval. We choose
AutoCut as the qualitative baseline because it is the most directly
comparable end-to-end method for e-commerce advertising video creation:
both systems consume product-related materials and output complete
videos. For each product, Fig.~\ref{fig:qualitative_comparison} shows
the product attributes (title, category, price, and customer feedback),
the shot-level narratives of AutoCut (top) and CutAgent (bottom), the
corresponding visual frames, and the narrative units realized by
CutAgent, together with the observed CTR of each video.

\noindent\textbf{Glass Teapot Set.}
In the glass teapot example (1000\,mL Glass Teapot Set; category: Tea
Set; price: 12.99 RMB), customer feedback emphasizes style, durability,
heat resistance, easy cleaning, and cost-effectiveness. AutoCut opens
with two highly similar shots that repeatedly present the product under
nearly identical viewpoints (0--3\,s), which wastes early attention and
reduces visual diversity. Subsequent shots mention a stainless-steel
filter and detachable accessories, but the overall storyline remains
close to a flat selling-point list. By contrast, CutAgent organizes the
advertisement as a conversion-oriented narrative sequence. It first
captures everyday friction with attention capture (\texttt{NU-D3}: messy
or inconvenient ceramic tea sets), then reveals the solution through
product recognition (\texttt{NU-C1}), demonstrates detachable design and
usage scenarios with feature-demonstration units (\texttt{NU-H9},
\texttt{NU-Z2}), and finally consolidates purchase confidence via trust
building (\texttt{NU-V4}: quality assurance and reviews). The resulting
video covers complementary aspects of the product rather than redundant
close-ups, and the observed CTR increases from 1.83\% (AutoCut) to
2.89\% (CutAgent).

\noindent\textbf{Sandy-Soil Honey Sweet Potatoes.}
In the sandy-soil honey sweet potatoes example (category: Sweet Potato;
price: 2.99 RMB / 2\,jin), customer feedback highlights sweetness and
delivery speed, while also mentioning variability in freshness and
weight. AutoCut begins with an appealing melted-center shot, but Shot~4
(``advantages at a glance'') uses a comparatively low-quality and less
informative frame, weakening mid-video engagement before the final call
to action. CutAgent instead follows a gene-conditioned progression:
strong visual impact for attention capture (\texttt{NU-P0}), origin
traceability for trust building (\texttt{NU-K3}: sandy-soil farming and
on-site harvesting), close-up and overview for product recognition
(\texttt{NU-C1}, \texttt{NU-Q7}), usage-scenario demonstration
(\texttt{NU-Z2}), and quality assurance (\texttt{NU-V4}). The retrieved
materials remain clearer and more semantically aligned with each
narrative unit. Correspondingly, the observed CTR rises from 1.27\%
(AutoCut) to 2.81\% (CutAgent).

\section{Prompt Templates}
\label{app:prompts}

Fig.~\ref{fig:prompts} presents the complete prompt templates used across the CutAgent pipeline. We describe each group below.

\noindent\textbf{(A) Creative-Gene Generation.}
This prompt is used by GeneAgent at both training time (to define the SFT target format) and inference time (to generate creative genes for new products). Given the multimodal product input $P_i$---textual attributes (category, SKU title, price, selling points, review summary) together with the main and detail product images---the prompt instructs GeneAgent to output a structured creative gene comprising an ordered narrative-unit sequence $\widetilde{S}_i$ drawn from the vocabulary $\mathcal{NU}$, along with production rules for script style ($\widetilde{r}_{s,i}$), voice-over ($\widetilde{r}_{v,i}$), transitions ($\widetilde{r}_{t,i}$), background music ($\widetilde{r}_{m,i}$), and sound effects ($\widetilde{r}_{f,i}$).

\noindent\textbf{(B) VideoAgent Prompts.}
\emph{B1 (Shot-Script Planning)} takes the generated creative gene $\widetilde{G}_i$, the multimodal product input $P_i$, and the retrieved failure--correction constraints $\mathcal{E}_{i}^{\mathrm{ret}}$ as context, and instructs VideoAgent to expand the gene into an executable shot-by-shot storyboard $\mathcal{B}_i$, specifying visual descriptions, voice-over text, duration, and production requirements for each shot.
\emph{B2 (Narration and Subtitle Adaptation)} takes the expanded shot scripts and the voice-over rule $\widetilde{r}_{v,i}$, and rewrites the scripts into TTS-ready narration with appropriate subtitle styles, adjusting length when clip duration constraints apply.

\noindent\textbf{(C) Reverse-Extraction Prompts for CutAgent-70K.}
These prompts recover creative genes from historical advertising videos to construct supervision targets.
\emph{C1 (Voice Attribute Analysis)} extracts vocal characteristics (gender, pitch, speed, expression) from the separated speech track.
\emph{C2 (BGM Style Analysis)} summarizes the background music style (tempo, energy, emotion, genre) from the separated music track.
\emph{C3 (Shot-Level Multimodal Analysis)} processes each detected shot to identify its visual content, oral content (voice-over transcript), and narrative purpose, using voice-over as the primary cue.
\emph{C4 (Creative Gene Reconstruction)} synthesizes the shot-level analyses, audio attributes from C1--C2, and the narrative-unit registry into a complete creative gene in the target schema.

\section{System Demonstration}
\label{app:system}

Beyond offline evaluation, we package CutAgent as an interactive
workbench for operators who create e-commerce advertising videos at
scale. As shown in Fig.~\ref{fig:system}, the interface follows the same
Plan--Execute--Self-Evolve pipeline as our method. On the left, users
load a multimodal product representation $P_i$ (main/detail images and
textual attributes) and invoke GeneAgent to generate a structured
creative gene (Fig.~\ref{fig:prompts}(A)). The center hub visualizes
narrative units and production rules in $\widetilde{G}_i$, allowing light
manual inspection before rendering. On the right, VideoAgent retrieves
category-then-gene constraints from $\mathcal{E}$, then executes
gene-guided shot planning, clip retrieval, TTS, BGM matching, and
timeline rendering (Fig.~\ref{fig:prompts}(B)). The bottom Reflection
panel reports the six evaluation dimensions (NQ, SPC, VSC, MQA, MSS, TB)
with temporally grounded diagnoses, so operators can accept automatic
revisions or intervene when needed; verified corrections are written
back to $\mathcal{E}$. The glass teapot example in the figure is the
same product analyzed in Fig.~\ref{fig:qualitative_comparison}.

\section{Related Work: Generative Video Models}
\label{app:gen_discussion}

Sec.~\ref{sec:related} reviews retrieval- and template-based pipelines
together with agentic editing systems. A second and rapidly moving line
of work generates the frames themselves instead of assembling existing
footage, and it is directly relevant to the production modules of
CutAgent: a shot whose retrieved material is missing or unusable cannot
be produced at all, and on-screen text has to be composed rather than
overlaid. We group representative methods of this line by the property
they contribute.

\subsection{Layout and Narrative-Controllable Generation}

Lay2Story~\cite{ma2025lay2story} conditions a diffusion transformer on
an explicit layout so that a multi-panel story keeps its subjects
consistent, and PlanGen~\cite{he2025plangen} goes further by planning
the spatial layout inside an auto-regressive vision--language model
before any pixel is generated. RelaCtrl~\cite{cao2026relactrl} shows
that control signals contribute very unevenly across layers and that
routing them by relevance keeps control cheap. These results are
consistent with the premise behind GeneAgent: structure should be
decided explicitly before content is produced, and it should be compact
enough to condition generation on. The two families differ mainly in the medium of the structure.
For these generators the structure is a latent layout condition consumed
by the generator, whereas for CutAgent the same prior decision is
written as an editable production plan---shot instructions, retrieval
queries, narration, subtitle and music rules---that can be scored
against a business objective and revised in place.

\subsection{Subject, Style, and Physical Consistency}

MoFu~\cite{ling2026mofu} preserves the scale and identity of several
reference subjects through scale-aware modulation and Fourier fusion,
and removes the sensitivity to reference ordering.
U-StyDiT~\cite{zhang2025ustydit} repaints ultra-high-resolution images
in an arbitrary artistic style with a diffusion transformer, while
WISA~\cite{wang2025wisa} injects explicit physical guidance so that a
text-to-video model respects physical laws instead of producing
plausible-looking motion. CustomTTT~\cite{bi2025customttt} customizes
appearance and motion jointly by test-time training, and
FancyVideo~\cite{feng2025fancyvideo} replaces frame-agnostic text
control with cross-frame textual guidance, which improves both dynamics
and temporal consistency over long durations. These methods target the
same failure mode: divergence of subject identity, style, or physical
behaviour as a synthesized video grows longer. Retrieval avoids most of
it by construction, since the advertised product is filmed rather than
re-imagined, but it pays in coverage---if the material for a required
shot is missing or unusable, the shot cannot be produced at all. These
generators are therefore the natural synthesizers behind VideoAgent's
shot plan (Sec.~\ref{app:videoagent}): the plan already names the
subject, the style, and the behaviour a shot needs, which is precisely
the conditioning these models consume. Drift also accumulates when no control signal
is involved at all: the multi-step samplers of a flow-matching backbone
amplify the discrepancy between the training and the inference
trajectory. DEFAR~\cite{huang2026exposure} turns this exposure bias into
its own correction signal, deriving directional and frequency-adaptive
feedback from a simulated single-step inference, and thereby raises the
model's tolerance to its own deviation. Rectification at this backbone
level complements the control-side methods above.

\subsection{Visual Text and Spatial Reasoning}

Advertising video carries text---selling points, prices, call-to-action
lines---whose legibility matters as much as the footage.
InnoText~\cite{liu2026innotext} unifies visual text generation and
editing in a single model and targets the structural regularity that
small and non-Latin glyphs demand, which is where the render-time
subtitle burn-in of Sec.~\ref{app:videoagent} is weakest; a shot-local
text renderer is a direct upgrade that leaves the rest of the pipeline
untouched. On the evaluation side, Lu et al.~\cite{lu2026space} find
that leading MLLMs fail textbook spatial-reasoning problems by a wide
margin, so a score produced by an MLLM evaluator is best read as a
relative signal between methods rather than an absolute measurement of
quality, which is why the metric suite of Sec.~\ref{app:metrics} is
always reported together with the forced-choice human study of
Sec.~\ref{app:human}.

\end{document}